\documentclass[letterpaper]{article} % DO NOT CHANGE THIS
\usepackage{aaai25}  % DO NOT CHANGE THIS[submission]
\usepackage{times}  % DO NOT CHANGE THIS
\usepackage{helvet}  % DO NOT CHANGE THIS
\usepackage{courier}  % DO NOT CHANGE THIS
\usepackage[hyphens]{url}  % DO NOT CHANGE THIS
\usepackage{graphicx} % DO NOT CHANGE THIS
\usepackage{natbib}  % DO NOT CHANGE THIS AND DO NOT ADD ANY OPTIONS TO IT
\usepackage{caption} % DO NOT CHANGE THIS AND DO NOT ADD ANY OPTIONS TO IT
\usepackage{algorithm}
\usepackage{algorithmic}

\usepackage{newfloat}
\usepackage{listings}
\DeclareCaptionStyle{ruled}{labelfont=normalfont,labelsep=colon,strut=off} % DO NOT CHANGE THIS
\floatstyle{ruled}
\newfloat{listing}{tb}{lst}{}
\floatname{listing}{Listing}
\usepackage{enumitem}
\usepackage{multirow}
\usepackage{makecell}
\usepackage{amsmath}
\usepackage{amsfonts}
\usepackage{stmaryrd}
\usepackage{booktabs}
\usepackage{tcolorbox}

\title{Test-time Controllable Image Generation by Explicit\\Spatial Constraint Enforcement}
\author {
    Zhexin Zhang\textsuperscript{\rm 1},
    Buyu Liu\textsuperscript{\rm 2},
    Jun Bao\textsuperscript{\rm 2},
    Suguo Zhu\textsuperscript{\rm 1},
    Long Chen\textsuperscript{\rm 3},
    Jun Yu\textsuperscript{\rm 2}
}
\affiliations {
    \textsuperscript{\rm 1}Hangzhou Dianzi University\\
    \textsuperscript{\rm 2}Harbin Institute of Technology (Shenzhen)\\
    \textsuperscript{\rm 3}The Hong Kong University of Science and Technology\\
}
\begin{document}
\maketitle

% \begin{abstract}
% AAAI creates proceedings, working notes, and technical reports directly from electronic source furnished by the authors. To ensure that all papers in the publication have a uniform appearance, authors must adhere to the following instructions.
% \end{abstract}

\begin{abstract}
Recent text-to-image generation favors various forms of spatial conditions, e.g., masks, bounding boxes, and key points. However, the majority of the prior art requires form-specific annotations to fine-tune the original model, leading to poor test-time generalizability. Meanwhile, existing training-free methods work well only with simplified prompts and spatial conditions. In this work, we propose a novel yet generic test-time controllable generation method that aims at natural text prompts and complex conditions. Specifically, we decouple spatial conditions into semantic and geometric conditions and then enforce their consistency during the image-generation process individually. As for the former, we target bridging the gap between the semantic condition and text prompts, as well as the gap between such condition and the attention map from diffusion models. To achieve this, we propose to first complete the prompt w.r.t. semantic condition, and then remove the negative impact of distracting prompt words by measuring their statistics in attention maps as well as distances in word space w.r.t. this condition. To further cope with the complex geometric conditions, we introduce a geometric transform module, in which Region-of-Interests will be identified in attention maps and further used to translate category-wise latents w.r.t. geometric condition. More importantly, we propose a diffusion-based latents-refill method to explicitly remove the impact of latents at the RoI, reducing the artifacts on generated images. 
% As for the former, we target bridging the gap between the semantic condition and that in images generated from natural text prompts, where we observe that there exist mismatches between the semantic condition and the semantic content in the generated scene. To achieve this, we propose to remove the negative impact of distracting prompt words by measuring their statistics in attention maps as well as distances in word space w.r.t. semantic condition. To further cope with the complex geometric conditions, we introduce a geometric transform module, in which Region-of-Interests will be identified in attention maps and later on, used to translate category-wise latents w.r.t. geometric condition. More importantly, we propose a diffusion-based latents-refill method to explicitly remove the impact of latents at RoI, reducing the artifacts on generated images. 
% Finally, negative prompts are incorporated to suppress hallucinated objects, which further improves the overall text-to-image generation performance. 
Experiments on Coco-stuff dataset showcase 30$\%$ relative boost compared to SOTA training-free methods on layout consistency evaluation metrics. Our code will be made available upon acceptance.

% To cope with the generic wrong association problem between the activation maps from the diffusion model and semantics from spatial constraint, our method removes the negative impact of distracting words by measuring their statistics as well as distances to semantics from spatial constraint. Once the target activation map is figured out, our method then handles single or multiple instances of the same semantic category separately given their different properties. As for the former, we first identify the Region-of-Interests of the single instance on the target attention map, then drive such a region w.r.t. spatial constraint. As for the latter, multiple instances of the same category are jointly considered. By further introducing a loss function that considers various forms , we are able to enforce more precise spatial constraint. We validate our ideas on 

% Specifically, our proposed method firstly figures out the association between the activations maps from diffusion model and semantics from spatial constraint at intermediate denoising steps, by explicitly removing the negative impact of distracting words. Then we drive the true activation regions to the target location. Finally, negative prompts are enforced to suppress hallucinated objects. Experiments on Coco-stuff dataset showcase 100% relative AP boost compared to SOTA training-free methods. Our code will be made available upon acceptance.
\end{abstract}

%Then multi-view perspective semantics are stitched together w.r.t. planer assumption to facilitate interactions between various views. Finally, our method processes stitched perspective semantics in parallel with a pre-trained text-to-image diffusion model, while integrating novel homographical loss to enforce global visual consistency. 


\section{Introduction}
\label{sec:intro}
Text-to-image generation has been a heated topic after the immersion of stable diffusion models~\cite{caesar2020nuscenes}. More recently, research focus has drifted to text-to-video generation, especially on generating long videos~\cite{bar2024lumiere}. Despite this, controllability remains one of the key challenges in generative artificial intelligence (AI) models where requests on detailed semantics, geometry, and relationships should be precisely presented in generated results. Typically, requests are of various modalities, ranging from sparse key points to dense pixel-level masks~\cite{kim2023dense}, aiming to allow fine-grained control to widen the applications of text-to-image generative models for more realistic scenarios. Achieving these requirements, 
%Once those coarse-to-fine conditions can be fulfilled, 
users can interactively cooperate with generative models, stimulating more potential for content creation~\cite{li2021image}.

\begin{figure}
    \centering %表示居中
    \includegraphics[width=8.2cm]{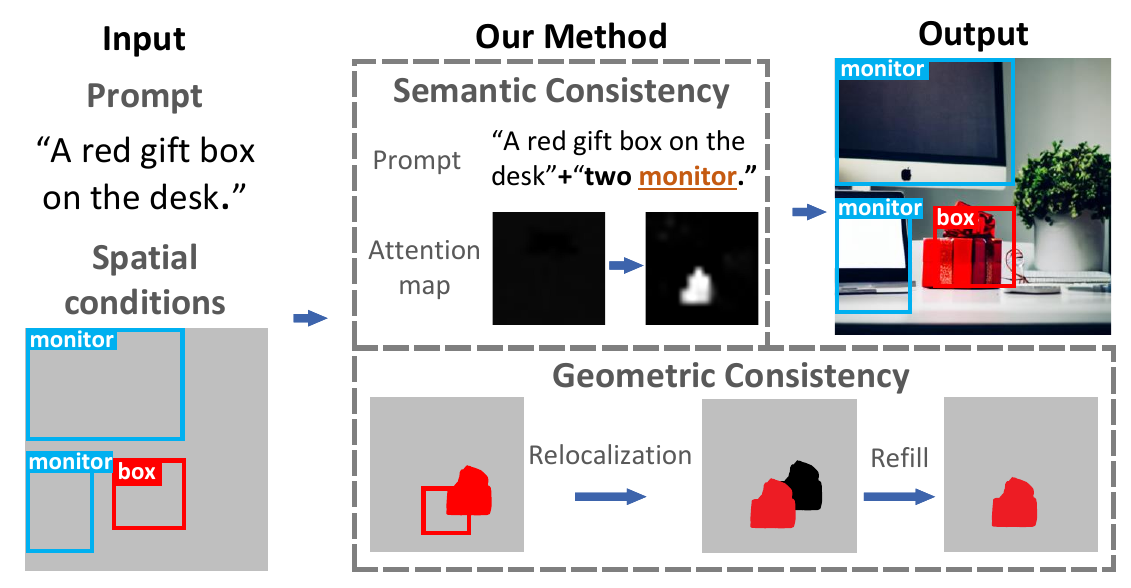} %
    % [height=4.5cm]表示高度
    %[width=9.5cm]表示宽度
    \caption{By enforcing semantic and geometric consistency, we can generate images with more precisely-located objects w.r.t. given layouts. 
    % By addressing semantic-mismatch problem and using geometric transformation, our method is able to generate images with correct layout.
    } %Overview of our method. 
    %图片的名称
    \label{fig_t}
    %图片的标签，用于文章中的引用，注意到标签的数字与实际文章显示的数字可能不同
\end{figure}

Scene layout, among the diverse conditions, plays an important role in literature as it provides a compact representation for semantic and geometric information~\cite{ashual2019specifying,kim2023dense}. Specifically, text-layout-to-image generation models~\cite{xie2023boxdiff,zheng2023layoutdiffusion,qu2023layoutllm} target at synthesizing images adhering to both the spatial conditioning input and the text prompt. However, previous work often relies on paired data, e.g., ground truth layout and original image, to either train or finetune models~\cite{solarte2022360}, which is less practical considering the time-consuming and labor-intensive ground truth labelling process. Moreover, introducing such data also limits the scalability of generative models as it is infeasible to extend to novel classes that are unseen in collected data. To address the above-mentioned bottlenecks, BoxDiff~\cite{xie2023boxdiff} proposes a training-free mechanism where special loss is introduced to activate and suppress attention maps w.r.t. provided spatial constraint. Though providing impressive results, we observe that BoxDiff suffers from simplified prompts, e.g., ``a $\{\}$ and a $\{\}$", and complex layouts where multiple instances of the same category are presented. For instance, spatial constraints with prompts like ``Three teddy bears, each a different color, snuggling together." would lead to unsatisfactory results where inconsistency happens both semantically and geometrically between generated images and prompts, or spatial constraints.

To this end, we propose a novel test-time controllable method that aims at natural languages as well as complex layouts, which can be found in Fig.~\ref{fig_t}. Specifically, we decouple the semantic and geometric conditions in spatial constraints and enforce their consistency individually. As for the former, we observe that there exists a mismatch between the word semantic in the prompt itself and the semantic content in the generated scene in the original stable diffusion models when natural text prompts are provided. For instance, given the prompt ``a person wearing a wet suit standing on a surfboard.", the latent map of the word ``suit" has significantly stronger relationships with the generated person in the scene than that of the word ``person" (See Fig.~\ref{3-2-att}). Not surprisingly, working on the latent map that is associated with word semantics in prompt would lead to unsatisfactory results. To this end, we propose to first address the mismatch problem by identifying the right word token w.r.t. semantic condition and then work on the right latent map. Specifically, given the semantics in spatial conditions, the right word token will be the semantics if its associated latent map is inactivated in the majority. Otherwise, we will rank all word tokens w.r.t. to their word distance to the semantic condition in ascending order, compute the statistics in their activation maps, and choose the first one that is regionally activated. By evaluating both statistics and word distance, we are able to remove the negative impact of unrelated word tokens, leading to better semantic consistency.

\begin{figure*}
    \centering %表示居中
    \includegraphics[width=0.89\linewidth]{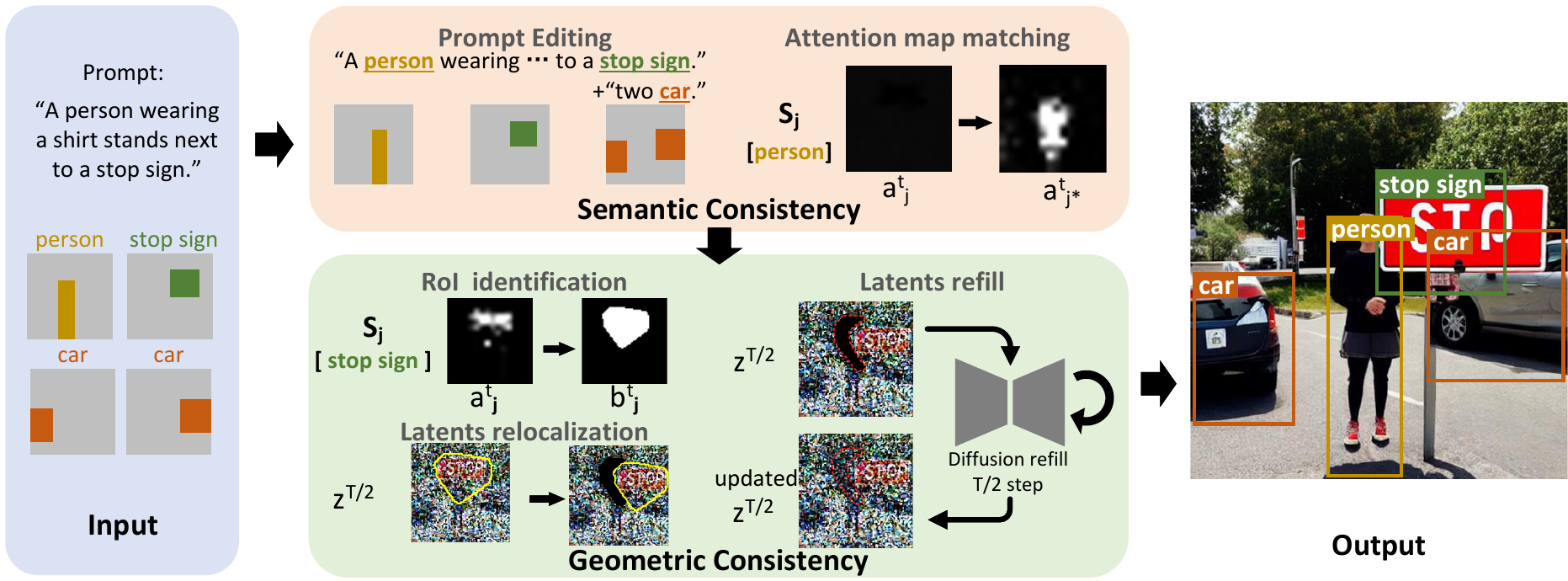} %height=6.5cm,  width=16cm
    % [height=4.5cm]表示高度
    %[width=9.5cm]表示宽度
    \caption{Given text prompts and layouts, our method enforces semantic consistency by prompt editing and attention map matching. Then the geometric consistency is incorporated by identifying, relocating, and refilling the latents. Thanks to both designs, our method leads to more realistic and consistent image generating process.
    % Our method consists of semantic part and geometric consistency part. In semantic consistency, the prompt will be edited to control the "car" during generation, and we rematched the attention map to control the "person". In geometric consistency, we propose the method of identification relocation and refill to adjust items to the correct position geometrically.
    }
    % By addressing semantic-mismatch and explicit geometric transformation problem, our method able to generates images with a more accurate layout.
    %图片的名称
    \label{main}
    %图片的标签，用于文章中的引用，注意到标签的数字与实际文章显示的数字可能不同
\end{figure*}

% better concentration on the right attention map. 
Once the right associated attention maps are obtained, our next step is to enforce the geometric consistency. Rather than activating and suppressing the instance-level attention maps based on geometric condition, which suffers heavily when multiple instances of the same category are presented or target location and the original activated region are spread out (See Fig.~\ref{main}(c)), we propose to explicitly figure out the activated regions categorically and then drive them to the target locations, in which Region-of-Interests will be identified in attention maps, utilized to segment out and translate latents w.r.t. geometric conditions. Moreover, we further propose a diffusion-based latent-refill method to adaptively remove the impact of latents at the original RoI.
% We further introduce negative prompts to suppress hallucinated objects. These explicit actuations allows more precise yet natural image generation.

Following prior art~\cite{zheng2023layoutdiffusion}, we validate our ideas on Coco-stuff dataset~\cite{caesar2018cocostuff} as it provides natural language descriptions with diverse layouts. Specifically, we report the coherency between the generated image and text prompt, as well as spatial conditions. Both the CLIP score (CS)~\cite{radford2021learning} and average precision (AP)~\cite{li2021image} are utilized to evaluate such coherency. In addition, FID~\cite{heusel2017fid} is also exploited as an evaluation metric for image quality. Compared to layout-to-image state-of-the-art methods under training-free category, our proposed method outperforms them under all evaluation metrics by 9$\%$ in a relative manner. Our contributions can be summarised as follows:
\begin{itemize}
    \item We introduce a novel test-time controllable text-to-image method that is generally applicable to natural language description and diverse spatial constraints.
    \item Our method decouples the spatial constraints into semantics and geometry. By addressing the semantic-mismatch and explicit geometric transformation problem, ours allows more natural yet precise image generation.
    \item Results on Coco-stuff demonstrate that ours outperforms SOTA methods by 9$\%$ under all evaluation metrics and by 30$\%$ in terms of layout consistency, relatively. Our code will be made available.
\end{itemize}

%Results on Coco-stuff demonstrate that ours outperforms SOTA methods under all evaluation metrics by 9$\%$ in a relative manner.  Our code will be made available.

\section{Related Work}
\label{sec:relatedwork}
\noindent{\textbf{Text-to-Image Generative Models}} The recent progress in text-to-image generation is primarily attributed to the generative networks pre-trained on large-scale data~\cite{reed2016generative,zhang2017stackgan}, thanks to the large-scale image-text pairs available on the Internet. Amidst a broad range of generative networks, diffusion models~\cite{ho2020denoising,song2020denoising} are being recognized as a promising family of generative models for generating image contents, by learning a progressive denoising process from the Gaussian noise distribution to the image distribution. Foundational work such as DALL-E~\cite{ramesh2021dalle}, GLIDE~\cite{nichol2021glide}, LDMs~\cite{rombach2022ldms}, and Imagen~\cite{saharia2022imagen} have showcased significant capabilities in text-conditioned image generation with higher quality and richer diversity. Besides text-to-image synthesis tasks~\cite{rombach2022high,li2023gligen}, diffusion models have proven exceptional across diverse tasks, such as inpainting~\cite{liu2018image}, and instructional image editing~\cite{kawar2023imagic}, due to their adaptability and competence in managing various forms of controls~\cite{zhang2023adding} and multiple conditions~\cite{qi2023layered}. Typically, those models are trained on extensive datasets and leverage the power of pre-trained language models. Our method aims to address the problem of effectively enforcing spatial conditions in text-to-image generation with re-training or fine-tuning, resulting in far better test-time generalizability.

\noindent{\textbf{Conditional Generative Models}}  Note that the above-mentioned text-based information roughly specifies the image contents, disallowing a fine-grained control of the image contents. To address the above problem, the multi-modal image information
like layout images~\cite{yang2022modeling}, semantic segmentation maps~\cite{qu2023layoutllm}, object sketches~\cite{kim2023dense}, and depth images~\cite{zhang2023adding} have been involved in hinting the image generation. Since the goal of our paper is to incorporate spatial conditions, we focus on the layout-to-image line of work. Prior to diffusion models~\cite{ho2020denoising}, adversarial networks (GAN) have dominated the image generation area. For example, a re-configurable layout is exploited in LostGANv2~\cite{sun2022LostGANv2} to better control over individual objects. Despite encouraging progress in this field, GAN-based methods suffer heavily from unstable convergence~\cite{ashual2019specifying}, mode collapse~\cite{johnson2018image}, and single modality of the control signal. Recently, fine-tuning stable diffusion models to adhere to additional layout information has also been explored in literature~\cite{kim2023dense,zheng2023layoutdiffusion,couairon2023zero}. However, they are trapped in a dilemma of time-consuming and labor-intensive annotation like box/mask-image paired data. Meanwhile, such data requirements also limit them to a closed-world assumption, failing to synthesize novel categories in the open
world. To this end, more recent work~\cite{xie2023boxdiff} proposes to operate constraints on the cross-attention to control the synthetic contents, avoiding additional training or close-world assumption. Compared to existing work, we propose a novel method that performs explicit manipulation on latents, resulting in more precise localization in a training-free manner.

% \begin{figure*}
%     \centering %表示居中
%     \includegraphics[width=16cm]{figure/method_fig/Main_v9.pdf} %height=6.5cm,
%     % [height=4.5cm]表示高度
%     %[width=9.5cm]表示宽度
%     \caption{Our method consists of semantic part and geometric consistency part. In semantic consistency, the prompt will be edited to control the "car" during generation, and we rematched the attention map to control the "person". In geometric consistency, we propose the method of identification relocation and refill to adjust items to the correct position geometrically.}
%     % By addressing semantic-mismatch and explicit geometric transformation problem, our method able to generates images with a more accurate layout.
%     %图片的名称
%     \label{main}
%     %图片的标签，用于文章中的引用，注意到标签的数字与实际文章显示的数字可能不同
% \end{figure*}

\section{Our Method}
\label{sec:method}
In this section, we provide more details about our method. Our method is based on stable diffusion models, which are briefly summarized in Sec.~\ref{sec:method_sd}. Then it decouples the semantic and geometric conditions from spatial ones, followed by enforcing the semantic consistency. Specifically, such consistency includes the consistency between prompts and semantic constraints, as well as the latter with attention maps from diffusion models (See Sec.~\ref{sec:method_semantic}). 
%right association between the semantic condition and attention map from diffusion models. 
Later on, the geometric consistency is explicitly incorporated by our geometric-transform module, leading to more precise location imposition in Sec.~\ref{sec:method_geometric}. We then provide more information about the inference procedure in Sec.~\ref{sec:method_inf}. 

Denoting the natural text prompt as $\textbf{l}=\{l_j\}_{j=1}^J$ consisting of $J$ words and the pre-trained stable diffusion model as $f$, the text-to-image process can be formulated as $I=f(\textbf{l})$, where $I\in\mathbb{R}^{H\times W\times 3}$ is the generated image. And we omit the input random variable $\textbf{x}$ for clarity. We further refer spatial condition as to $\textbf{c}=\{\textbf{s},\textbf{g}\}$, composing of semantic condition $\textbf{s}=\{s_n\}_n$ and geometric condition $\textbf{g}=\{g_n\}_n$. Specifically, $s_n$ and $g_n$ are the corresponding conditions for the target object or stuff $n=\{1,\dots,N\}$. Typically, $s_n$ represents its semantic class and $g_n$ can be either a pair of a set of paired $x,y$ locations, depending on what type of spatial condition is provided. Later in our paper, we will follow the popular setup~\cite{zheng2023layoutdiffusion,xie2023boxdiff} where bounding boxes are used as $\textbf{c}$. And our method is not limited to bounding box but can be applicable to diverse spatial conditions (See Sec.~\ref{sec:experiment}). 
% Please note that our method is not limited to exploiting bounding boxes. For extensions to more diverse spatial conditions, we refer the readers to Sec.~\ref{sec:experiment}.
% We assume there exist $N$ targets in the current setting, and $s_n$ and $g_n$ are the corresponding conditions for the target $n\in\{1,\dots,N\}$. 
% The ultimate goal of this paper 
Our goal is to generate an image $I=f(\textbf{l},\textbf{c})$ that meets the requirements of both $\textbf{l}$ and $\textbf{c}$, without re-training or finetuning the model $f$ (See Fig.~\ref{main}).

% Conventionally, $s_n$ represents its semantic class and $g_n$ can be either a pair of a set of paired $x,y$ locations, depending on what type of spatial condition is provided. Later in our paper, we will follow the popular setup~\cite{??} where bounding boxes are used as $\textbf{c}$. Please note that our method is not limited to exploiting bounding boxes. For extensions to more diverse spatial conditions, we refer the readers to Sec.~\ref{sec:experiment}.

\subsection{Preliminaries: Stable Diffusion Models}\label{sec:method_sd}
Here we provide some background for the stable diffusion model $f$. In general, diffusion models are probabilistic models designed to learn a data distribution $p(\textbf{x})$ by gradually denoising a normally distributed variable $\textbf{x}$, which corresponds to learning the reverse process of a fixed Markov Chain of length $T$~\cite{ho2020denoising,song2020denoising}. More than often, diffusion models operate on latent space with the help of encoder $\mathcal{E}$ and decoder $\mathcal{D}$. Given $\textbf{x}$, the encoder $\mathcal{E}$ encodes it into a latent representation $\textbf{z}=\mathcal{E}(\textbf{x})$, and the decoder $\mathcal{D}$ reconstructs an image back from the latent, resulting in $\hat{\textbf{x}}=\mathcal{D}(\textbf{z})=\mathcal{D}(\mathcal{E}(\textbf{x}))$. At each time step $t\in\{0,\dots,T\}$, a noisy latent $\textbf{z}_t\in\mathbb{R}^{h\times w\times c}$ will be obtained, where $h$, $w$ and $c$ are the height, width and channel dimension, respectively. 
% In practice, the latent $\textbf{z}\in\mathbb{R}^{h\times w\times c}$. Importantly, the encoder downsamples the image by a factor $r=\frac{H}{h}=\frac{W}{w}$, where $r$ is often defined as an even number.

Compared to diffusion models, stable diffusion model $f$ further allows additional control, such as text prompt $\textbf{l}$, to guide the image content. Such additional control is often pre-processed and converted to tokens by off-the-shelf encoders~\cite{rombach2022high}. Denoting the text encoder as $\tau_{\theta}$, the training process of $f$ can be defined as:
\begin{equation}
    \mathcal{L}_f := \mathbb{E}_{\textbf{z}\sim\mathcal{E}(\textbf{x}),\textbf{l},t,\epsilon\sim\mathcal{N}(0,1)}[\|\epsilon-\epsilon_{\theta}(\textbf{z}_t,t,\tau_{\theta}(\textbf{l}))\|_2^2]
\end{equation}
where the neural backbone $\epsilon_{\theta}$ of stable diffusion model is often realized as a time-conditional UNet~\cite{ronneberger2015unet}. During inference, a latent $\textbf{z}_T$ is sampled from the standard normal distribution $\mathcal{N}(0,1)$ and the $f$ is used to iteratively remove the noise in $\textbf{z}_T$ to produce $\textbf{z}_0$. Then the latent $\textbf{z}_0$ is passed to the decoder $\mathcal{D}$ to generate $I$. %generate an image $I$.

\subsection{Semantic Consistency}\label{sec:method_semantic}
After decomposing $\textbf{c}$ into $\textbf{s}$ and $\textbf{g}$, our first step is to enforce semantic consistency w.r.t. $\textbf{s}$. 

\begin{figure}[htbp]
    \centering %表示居中
    \includegraphics[width=\columnwidth]{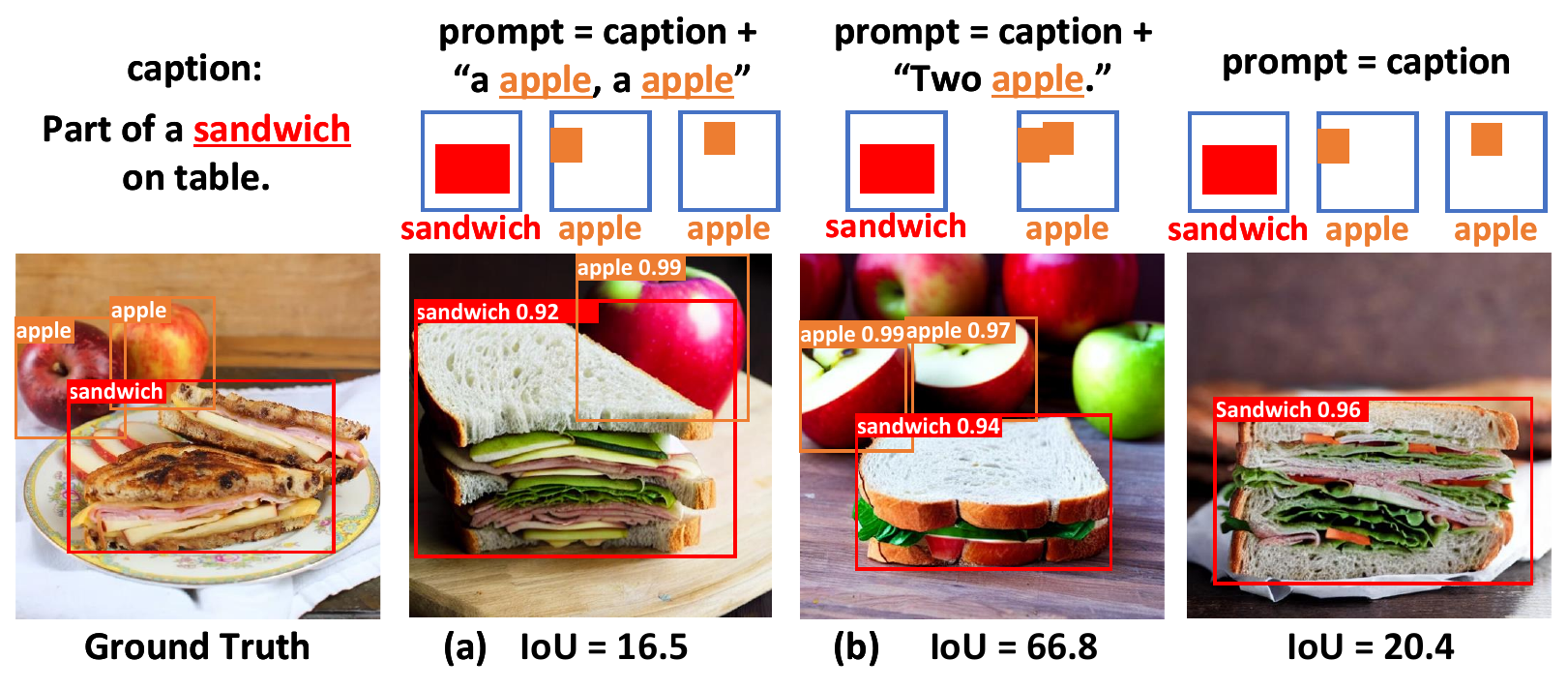} %
    \caption{Prompt editing enables the discovery of missing objects by comparing semantics in caption and layouts. 
    %to generate the object "apple"(missing in caption) correctly.
    }
    \label{prompt_editing}
\end{figure}

%\paragraph{Prompt editing} 
\noindent{\textbf{Prompt editing}} Ideally, the prompt $\textbf{l}$ is semantically consistent with $\textbf{s}$. In other words, $\textbf{s}\subset\textbf{l}$ 
% $\textbf{s}$ should be a subset of $\textbf{l}$, or $\textbf{s}\subset\textbf{l}$, 
as natural descriptions $\textbf{l}$ might cover high-level abstractions compared to pure semantics $\textbf{s}$. 
However, it happens that $s_n\not\in\textbf{l}$ in real-world dataset. Let's take an image from the Coco-stuff dataset for example~\cite{caesar2018cocostuff}. As can be found in Fig.~\ref{prompt_editing}.a, ``apple" is labelled as the category of the highlighted bounding box but is missing in caption ``Part of a sandwich on table". In this case, there is no way to recover object class ``apple" in $I$ with the current prompt.
% In this case, there is no way to generate $I$ with the object label "XX" at present with the current prompt.
To solve this semantic inconsistency problem, we propose to complete $\textbf{l}$ according to $\textbf{s}$, if necessary. 
% we propose to check whether $\textbf{l}$ is consistent w.r.t. $\textbf{s}$. And then revise the prompt if needed. 
Specifically, we first identify unique $s_n$s in $\textbf{s}$ and obtain their quantities, resulting in a new set $\{\{S_m,X_m\}\}_{m=1}^M$ where $S_m\in\textbf{s}$ and $X_m$ is the $m$-th unique semantic and its quantity. And $M$ is the number of unique semantics. For any semantics $S_m$ that are not included in $\textbf{l}$, we will include a new sentence '$X_m S_m$' to $\textbf{l}$. To this end, we are able to recover missing instances in prompts, leading to better generation results (See Fig.~\ref{prompt_editing}.b). 
% Examples can be found in Fig.~\ref{prompt_editing}.b.
% We refer the readers to Fig.~\ref{?}.b for example.   
% Please note that Boxdiff~\cite{??} use a much 

\begin{figure}[htbp]
    \centering %表示居中
    \includegraphics[width=0.95\linewidth]{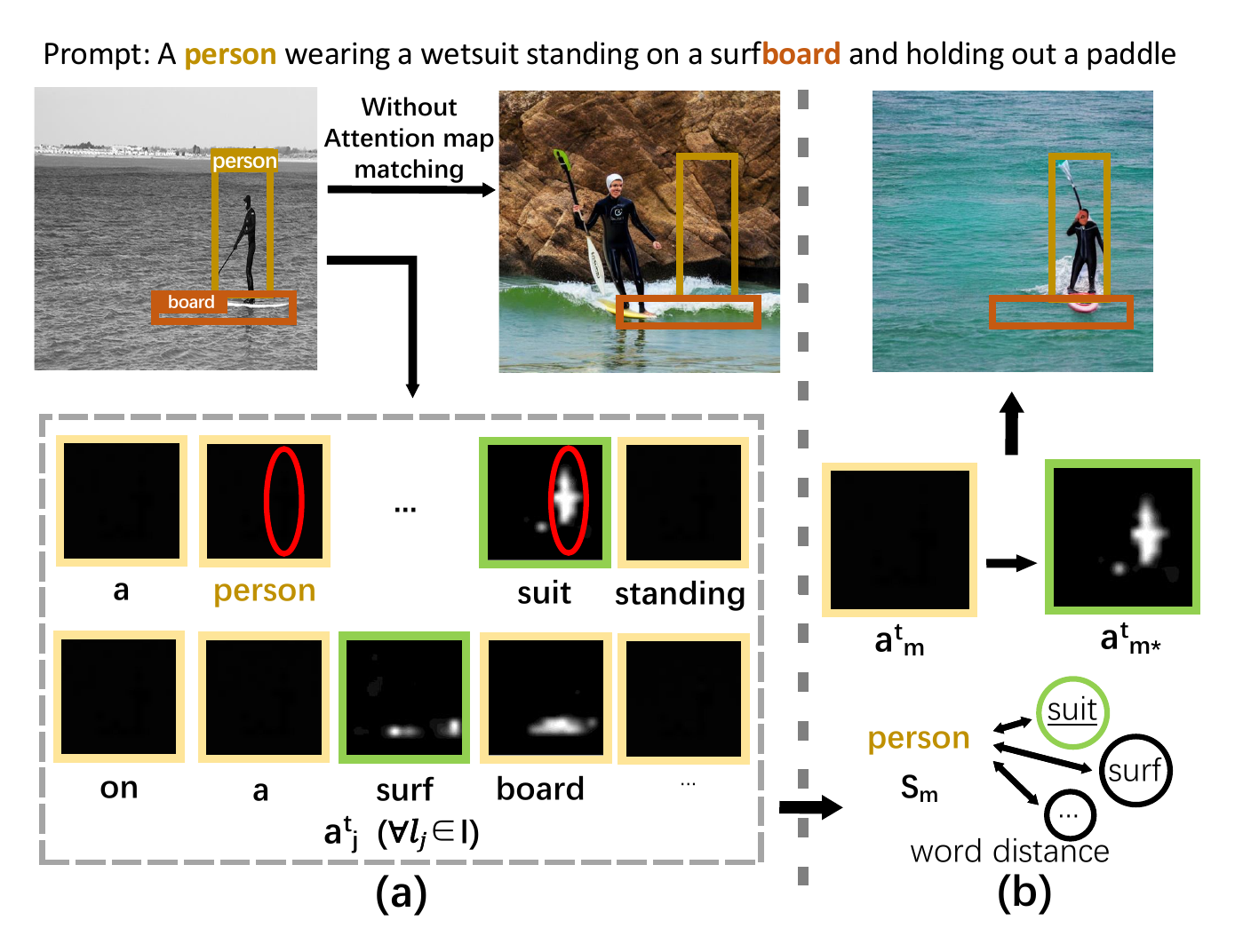} %width=\columnwidth width=8.4cm
    \caption{Instead of working on attention maps according to their default semantic token, we propose to match attention map w.r.t. their statistics and distance in word space.
    % We re-match tokens for the mis-matched object to confirm that the controlled token is correct.
    }
    \label{3-2-att}
\end{figure}

\noindent{\textbf{Attention map matching}} After editting the prompt, we can obtain a complete $\textbf{l}$ that is semantically consistent with $\textbf{s}$. Our next step is to identify the corresponding attention map for each $S_m$ for more accurate manipulation. Though this is not a problem under simplified prompt, e.g., the 'a $s_n$' style prompt used in Boxdiff~\cite{xie2023boxdiff}, we observe that mismatch happens under naturally described prompt $\textbf{l}$, leading to wrong actuation at later stages. 
% We provide an example of a natural language prompt, the image generated with $f$, and the visualized attention maps in Fig.~\ref{3-2-att}.a. 
% Clearly, 
As can be found in Fig.~\ref{3-2-att}.a, the latent map of 'suit' has a more visible or closer relationship with the generated person (highlighted by the red circle) than that of 'person'.

To this end, we propose to check the statistics of $l_j$ in the attention map as well as compute its word distance to $S_m$. 
In practice, $\forall l_j\in\textbf{l}$, we collect the statistics on the attention map of each $l_j$, or $a_j^t$, in order and decide whether this map is partially activated. 
Later on, we compute their distance to $S_m$ in word-vector space~\cite{radford2021learning}. 
Then the words with partially activated map are ranked based on the distance score in ascending order, and we select the first one.
%In practice, $\forall l_j\in\textbf{l}$, we compute its distance to $S_m$ in word-vector space~\cite{radford2021learning}. Then all $J$ words are ranked based on the distance score in ascending order. 
%%And this is the order in which the later statistics are collected. 
%Later on, we collect the statistics on the attention map of each $l_j$, or $a_j^t$, in order and decide whether this map is partially activated, %e.g., XXXX. % and then stop the process 
%We stop this process until the first partially activated map is found. 
We further denote the $a_{m^*}^t$ as the matched attention map for $S_m$. By figuring out $a_{m^*}^t$ for each $S_m$, our method is less prone to useless manipulations when leveraging geometric consistency at later stage. Examples of our matching results can be found in Fig.~\ref{3-2-att}.b. 
% Therefore, we are more likely to find the right latent for each $S_m$. Examples of our matching results can be found in Fig.~\ref{?}.b. 

\subsection{Geometric Consistency}\label{sec:method_geometric}
Compared to semantic consistency which aims to complete the prompt and find the right corresponding attention map, our geometric consistency targets precisely locating the generated instances to the right place, or according to $\textbf{g}$.

Though enforcing instance-level losses on attention map sheds some light on geometric consistency~\cite{xie2023boxdiff}, such a procedure relies heavily on the weights that control the activation and suppression on attention map, leading to unstable localization results. To this end, we propose a novel geometric transformation module that explicitly performs manipulations on latents. Our module is able to not only identify and relocate the Region-of-Interests (RoI) w.r.t. $g_n$ in latents, but also refill the RoI with reasonable values. % More details can be found in the following paragraphs.  

\noindent{\textbf{Geometric transformation module}} Our geometric transformation module consists of three main steps, including identifying RoI for each $s_n$ w.r.t. its attention map, relocating the corresponding area in latent to the target location $g_n$, and refilling the values in latents at the original RoI. 
% including identifying RoI for each $s_n$ w.r.t. its attention map, moving the corresponding area in latent to the target location $g_n$, and then re-filling the values in the original RoI. 

% \begin{figure}[htbp]
%     \centering %表示居中
%     \includegraphics[width=8.4cm]{figure/method_fig/RoI_identification_v6.pdf} %height=6cm,
%     \caption{We collected attention map and mask the object area, then convert the mask area to a convex hull.}
%     \label{RoI_identification}
% \end{figure}

\noindent{\textit{RoI identification}} aims to identify the RoI on the attention map associated with $s_n$. We concatenate all attention maps $\{a_j^t\}_j$ together and then normalize it to $[0,1]$ w.r.t. all $J$ words. Then we resize the normalized attention maps to the size of latent, or $\hat{\textbf{a}}^t=\in\mathbb{R}^{h \times w \times J}$.
% Then for the object $s_n$, Denoting the attention map that is associated with $s_n$ at the $t$-th time step as $a_t^n$, we resize it to the size of latent, or $h \times w$, and normalize the $a_n^t$ to $[0,1]$. 
Similarly, the ${n^*}$-th layer in $\hat{\textbf{a}}^t$ is the normalized and resized attention map that matches $s_n$. And the initial set of RoI is defined as groups of pixels whose values are greater than $\lambda$ in the the ${n^*}$-th layer. 
% Then regions in which the ${n^*}$-th layer  is greater than $\lambda$ are taken as the initial set of RoI. 
Intuitively, these activated pixels indicate the places where $s_n$ is generated by the stable diffusion model $f$. As we make no assumption about the size of $s_n$, 
% Since the size of $s_n$ is unknown beforehand, we make no assumption about it. Therefore, 
we choose a conservative pathway where a convex hull from these initial pixels is generated and presented by a binary mask $b_n^t\in\mathbb{R}^{h \times w}$. Pixel value is assigned to 1 if it is inside the hull. 
% Pixels that are inside the hull will be assigned the value 1 and otherwise are all 0. 
We refer to this binary mask as the indicator of our final RoI of $s_n$. %The entire process of RoI identification is visualized in Fig.~\ref{RoI_identification}.

\noindent{\textit{Latents relocalization}} Thanks to $b_n^t$ that indicates the RoI, which explicitly provides the location information of $s_n$ at the $t$-th time step, we can utilize it as a reference for latent manipulation. Let's denote the center of $g_n$ and $b_n^t$ as $c_n^g\in\mathbb{R}^2$ and $c_n^b\in\mathbb{R}^2$. Mathematically, we have:
\begin{equation}
\begin{split}
    &\textbf{z}^t  \llbracket [x_0,y_0]+c_n^g-c_n^b\rrbracket=\textbf{z}^t \llbracket [x_0,y_0] \rrbracket; \\ & \forall b_n^t\llbracket [x_0,y_0] \rrbracket==1 \ \  \&\& \ \ [x_0,y_0]+c_n^g-c_n^b\in\mathbb{R}^{h\times w} %\quad
\end{split}\label{eq:geo_trans}
\end{equation}
where $x_0\in\{1,\dots,h\}$ and $y_0\in\{1,\dots,w\}$ are the x,y locations in $b_n^t$. $b_n^t \llbracket [x_0,y_0] \rrbracket$ takes the value from $[x_0,y_0]$ in $b_n^t$. In practice, Equ.~\ref{eq:geo_trans} defines a translation such that the activated latents 
%that have been activated by $b_n^t$ 
will be copied to the target location, as long as they are in range.

% \begin{figure}[htbp]
%     \centering %表示居中
%     \includegraphics[width=8.3cm]{figure/method_fig/RoI_refill_Vis_v3_2.pdf} %height=6cm,
%     \caption{Different methods in RoI refill. Our method(c) has fewer artifacts than using binary image dilation(b).}
%     \label{RoI_refill_image}
% \end{figure}

\noindent{\textit{Latents refill}} Though latents relocalization explicitly copies the activated latents according to $g_n$, the problem of removing the negative impact of latents at $b_n^t$ remains. Therefore, our next step is to refill the these latents with natural values.
% Then the question becomes how to refill the these latents with natural values.

Conventionally, techniques of image process, such as binary image dilation~\cite{dhandra2006skew}, or computer vision, e.g., image inpainting~\cite{lugmayr2022repaint}, %can be utilized
are designed to refill areas w.r.t. their surrounding area photometrically in either latent or image space. The former, however, leads to artifacts in the complex de-noising process. While the latter is more of a post-processing step, thus is against the nature of the generative models. Instead, we introduce a diffusion-based technique that learns to refill the latents. %(See Fig.~\ref{RoI_refill_image})

\begin{figure}[htbp]
    \centering %表示居中
    \includegraphics[width=0.88\linewidth]{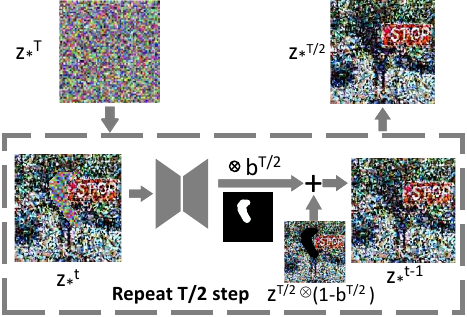} %height=6cm, width=7.7cm
    \caption{To refill $b^t$ with natural values, we introduce a lightweight diffusion model to iteratively update it. 
    % To refill the missing area $b^t$ caused by latents relocalization we use a T/2-step diffusion to generate it.
    }
    \label{RoI_refill}
\end{figure}

Denoting $\textbf{b}^t=\cup\{b_n^t\}_n$ as the union of the set of binary mask $b_n^t$, a naive way to refill $\textbf{z}$ at $\textbf{b}^t$ is to send a new noise to $f$, obtain the new latents $\textbf{z}_*^t$, and then replace the $\textbf{z}_t$ at $\textbf{b}^t$ with that of $\textbf{z}_*^t$ by:
\begin{equation}
  \textbf{z}^t \llbracket [x_0,y_0] \rrbracket = \textbf{z}_*^t \llbracket [x_0,y_0] \rrbracket, \quad \forall \textbf{b}^t\llbracket [x_0,y_0]\rrbracket == 1 ~\label{eq:latent_update}
\end{equation}
Apparently, such a naive combination may lead to artifacts at the boundary of $\textbf{b}^t$ as $\textbf{z}_*^t$ shares no information with $\textbf{z}^t$ during denoising. To solve this, we propose to update $\textbf{z}_*^t$ w.r.t. $\textbf{z}^t$. In practice, to obtain $\textbf{z}_*^t$, we first generate $\textbf{z}_*^T$ with a random noise. Then we update $\textbf{z}_*^T$ with the following equation:
\begin{equation}
    \textbf{z}_*^t = \textbf{z}_*^t \otimes \textbf{b}^t + \textbf{z}^t \otimes (1-\textbf{b}^t) ~\label{eq:sec_latent}
\end{equation}
And we iterate between Equ.~\ref{eq:sec_latent} and Equ.~\ref{eq:latent_update} until $0$ time steps. In practice, we found such an iterative method can be time-consuming and unnecessary. Therefore, we perform the RoI identification and latents relocation at time step $\frac{T}{2}$ only. And we obtain a simplified version of $\textbf{z}_*^{\frac{T}{2}}$ with:
\begin{equation}
    \textbf{z}_*^t = \textbf{z}_*^t \otimes \textbf{b}^{\frac{T}{2}} + \textbf{z}^{\frac{T}{2}} \otimes (1-\textbf{b}^{\frac{T}{2}}) \quad \forall t \geq \frac{T}{2}
\end{equation}
Later on, we apply Equ.~\ref{eq:latent_update} and de-noise the updated $\textbf{z}^t$ until $t$ equals to $0$. We visualize the latents refill process in Fig.~\ref{RoI_refill}. 

% are not only not well aligned with our generative nature, but also may lead to artifacts in the complex de-noising process (See Fig.~\ref{??} as a reference).

\subsection{Inference}\label{sec:method_inf}
During the denoising step of the stable diffusion model $f$, or inference step, the high response regions in the attention map are highly related to $s_n$ in the decoded image $I$. Therefore, we share a similar motivation of~\cite{xie2023boxdiff} and introduce losses on the attention maps to further enforce the spatial constraints $\textbf{c}$ of the synthesis of our targets. However, their loss function suffers when multiple objects of the same category are present, where the instance-level Inner-Box constraint and Outer-Box constraint fight against each other. Therefore, we propose a category-level loss for each unique $S_m$. By enforcing Inner-Box constraint and Outer-Box constraint on each category, we are able to guarantee more stable outputs compared to BoxDiff~\cite{xie2023boxdiff}. The overview of our method can be found in Fig.~\ref{main}.

% \paragraph{Categorical loss} 

\section{Experiment}
\label{sec:experiment}
% In this section, we will provide more details about our experiments, including the dataset, baselines, implementation details and results. 

% \subsection{Experimental Setup}
\noindent{\textbf{Dataset}}
Inspired by~\cite{zheng2023layoutdiffusion}, we use Coco-stuff~\cite{caesar2018cocostuff} dataset to validate our ideas where human annotations are provided for 80 object categories and 91 stuff categories. Specifically, we utilize COCO 2017 Stuff Segmentation Challenge subset, containing 40K/5k/5k images for train/val/test-dev set, respectively. To better exploit the layout-to-image generation process, we only select images with 3 to 8 objects that cover more than 2$\%$ of the image area and not belong to crowd, leading to 25,210 train and 3,097 val images. Unlike methods that require the massive data for training, ours is training-free thus can be directly applied on validation images. There are 5 human annotated captions for each validation image. The longest one is regarded as the prompt $\textbf{l}$.
% There exists 5 human annotated captions for each validation image. Among them, the longest caption is regarded as the prompt $\textbf{l}$. 

% \noindent{\textbf{Baselines and Evaluation Metrics}} See below the baselines and evaluation metrics we use in our experiment.
% We choose the following baselines and evaluation metrics in our experiments. 
\noindent{\textbf{Baselines}} We choose three main baselines both for reproductivity and performance purposes. \textbf{Stable Diffusion}~\cite{rombach2022ldms} is the generic generative model without any control signal. In constrast, \textbf{Layoutdiffusion}~\cite{zheng2023layoutdiffusion} is a fine-tuning based method that is trained on the 25210 Coco-stuff images. \textbf{BoxDiff}~\cite{xie2023boxdiff} is the first training-free method. In experiments, the authors validate their ideas on simplified prompts, e.g., ``a $s_n$", and easy layouts, e.g., at most two objects in image. %in their provided json file. 
%Similarly, we use their official implementation to generate results on natural yet more complex Coco-stuff dataset. 
We further simplify Coco caption with the format that designed by~\cite{xie2023boxdiff}. We refer it as ``simplified prompt". Unless otherwise notified, we use their official implementations. More details can be found in implementation details.

\begin{table*}[htbp]
    \begin{minipage}[t]{0.6\textwidth}

\fontsize{10}{13}\selectfont
\centering
\begin{tabular}{p{0.75cm}p{1cm}|p{0.6cm}|p{0.6cm}p{0.6cm}p{0.6cm}|p{0.63cm}p{0.63cm}p{0.63cm}}%{cc|c|cccccc}
\toprule
\multicolumn{2}{c|}{\multirow{2}{*}{Method}}  & \centering \multirow{2}{*}{\begin{tabular}[c]{@{}c@{}}Zero\\Shot\end{tabular}} & \multicolumn{3}{c|}{Layout Consistency} & \multicolumn{3}{c}{Image Quality} \\ \cline{4-9} 
\multicolumn{2}{c|}{}                         &                                               & \centering AP          & \centering AP50        & \centering AP75       & \centering FID       & \centering IS        & \multicolumn{1}{c}{\centering CS }       \\  \midrule
\multicolumn{2}{c|}{Stable Diffusion}         & \centering $\checkmark$                           & \centering 0.7         & \centering 2.4         & \centering 0.4        & \centering 36.05     & \centering -         & \multicolumn{1}{c}{\centering 26.81 }    \\ \hline
\multicolumn{2}{c|}{Layout Diffusion}         & \centering $\times$                      & \centering 32.0        & \centering -           & \centering -          & \centering 15.63     & \centering 28.36     & \multicolumn{1}{c}{\centering - }        \\ \hline
\multirow{2}{*}{Boxdiff} & \centering (a)  & \centering \multirow{2}{*}{$\checkmark$}          & \centering 0.9         & \centering 2.8         & \centering 0.4        & \centering 60.75     & \centering -         & \multicolumn{1}{c}{\centering 24.11 }    \\
                         & \centering (b)       &                                               & \centering 2.7         & \centering 8.0         & \centering 1.5        & \centering 33.57     & \centering -         & \multicolumn{1}{c}{\centering 26.30 }    \\ \hline
\multicolumn{2}{c|}{Ours}                     & \centering $\checkmark$                           & \centering 3.5         & \centering 9.7         & \centering 1.7        & \centering 39.86     & \centering -         & \multicolumn{1}{c}{\centering 24.69 }    \\ \bottomrule
\end{tabular}

% (a) simplified
% (b) COCO cap
%\centering \begin{tabular}[c]{@{}c@{}}\end{tabular} \centering \multirow{2}{*}{\begin{tabular}[c]{@{}c@{}} \fontsize{10}{10}\selectfont{Requires}\\ \fontsize{10}{10}\selectfont{Training}
    \caption{ Comparing with SOTA Zero-shot method Boxdiff, our method have better performance in the task of layout consistency. (a) Use Boxdiff simplified prompt. (b) Use COCO Caption as prompt. }
    \label{tab:main}
    \end{minipage}
  % \centering
    \begin{minipage}[t]{0.4\textwidth}
    % \begin{table}[]
\begin{tabular}{>{\centering}p{0.55cm}>{\centering}p{0.55cm}|>{\centering}p{0.4cm}>{\centering}p{0.5cm}>{\centering}p{0.5cm}>{\centering}p{0.65cm}c}
    \toprule
     Sem & Geo & AP & AP50 & AP75 & CS & FID \\ \midrule
    ×    &     ×     &  2.4  &  6.8    &  1.2    &    24.72    &  38.95   \\
    \checkmark    &     ×     &  3.1  &  9.2    &  1.4    &    26.30        &  38.37   \\
    ×    &     \checkmark     &  3.1  &  8.7    &  1.8    &    24.82        &  39.31   \\
    \checkmark    &     \checkmark     &  3.5  &  9.7    &  1.7    &    24.69        &  39.86   \\ \bottomrule
\end{tabular}
% \end{table}
%\makecell{\fontsize{9}{10}\selectfont{Sem}} \makecell{\fontsize{9}{10}\selectfont{Geo}}
    \caption{Results of ablation studies. Experimental results show the effectiveness of both Semantic Consistency (Sem) and Geometric Consistency (Geo).}
    % Sem refers to the Semantic Consistency and Geo refers to the Geometric Consistency of our method.
    \label{tab:ablation}
    \end{minipage}
\end{table*}

% \begin{table*}[htbp]
%   \centering
%   \input{table/main_table}
%   \caption{Performance on Coco-stuff dataset. Comparing with SOTA Zero-shot method Boxdiff, our method have better performance in the task of layout consistency. }
%   \label{tab:main}
% \end{table*}

\noindent{\textbf{Evaluation Metrics}} To measure the quality of generated images, we report the overall performance with metrics about layout consistency and image quality ~\cite{zheng2023layoutdiffusion}. % two sets of metrics %\textbf{Layout Consistency}
Layout Consistency measures whether generated images follow the spatial constraints. As for generic bounding box setup, Average Precision (\textbf{AP})~\cite{li2021image} is wide exploited in literature. We follow the conventions to report AP, AP50 and AP75 in our experiments. %D_01
Image Quality is measured by Fréchet Inception Distance (\textbf{FID})~\cite{heusel2017fid}, Inception Score (\textbf{IS})~\cite{salimans2016is}, and CLIP Score (\textbf{CS})~\cite{radford2021learning}. More details about these metrics can be found in our supplementary.
% The evaluation metrics will be detailed in part 3 of our supplementary.%D_02

% To measure the quality of generated images, we report the overall performance with the following two sets of metrics~\cite{zheng2023layoutdiffusion}:
% \begin{itemize}
%     \item \textbf{Layout Consistency} measures whether generated images follow the spatial constraints. As for generic bounding box setup, Average Precision (AP)~\cite{li2021image} is wide exploited in literature. We follow the conventions to report AP, AP50 and AP75 in our experiments.%D_01
%     \item \textbf{Image Quality} is measured by Fréchet Inception Distance (FID)~\cite{heusel2017fid}, Inception Score (IS)~\cite{salimans2016is}, and CLIP Score (CS)~\cite{radford2021learning}.%D_02
% \end{itemize}

\noindent{\textbf{Implementation Details}}
We have implemented the system with PyTorch~\cite{paszke2019pytorch} while using publicly available Stable Diffusion code~\cite{rombach2022ldms}. In experiments, we use a machine with GeForce RTX3090 GPU for inference. On average, it takes about 20s to generate an image with both prompts and spatial constraints. As a reference, it takes~\cite{xie2023boxdiff} 17s to perform a similar task, meaning that ours is equally efficient as SOTA method. $T$ is set to 50 for fair comparisons in all experiments. As for~\cite{xie2023boxdiff}, we revise their official implementation where they hard-coded the second or every third word in prompt as the corresponding semantics in single or multiple object setup. To fit the natural language in real Coco-stuff dataset, we choose the one when $l_j$ equals to $s_n$. In addition, we conduct a new experiment where the prompts are all generated w.r.t. spatial conditions in the ``a $s_n$" format, as suggested in their main paper. We refer the readers to supplementary for full implementation details.

\begin{figure}[t]
    \centering
    \includegraphics[width=0.93\columnwidth]{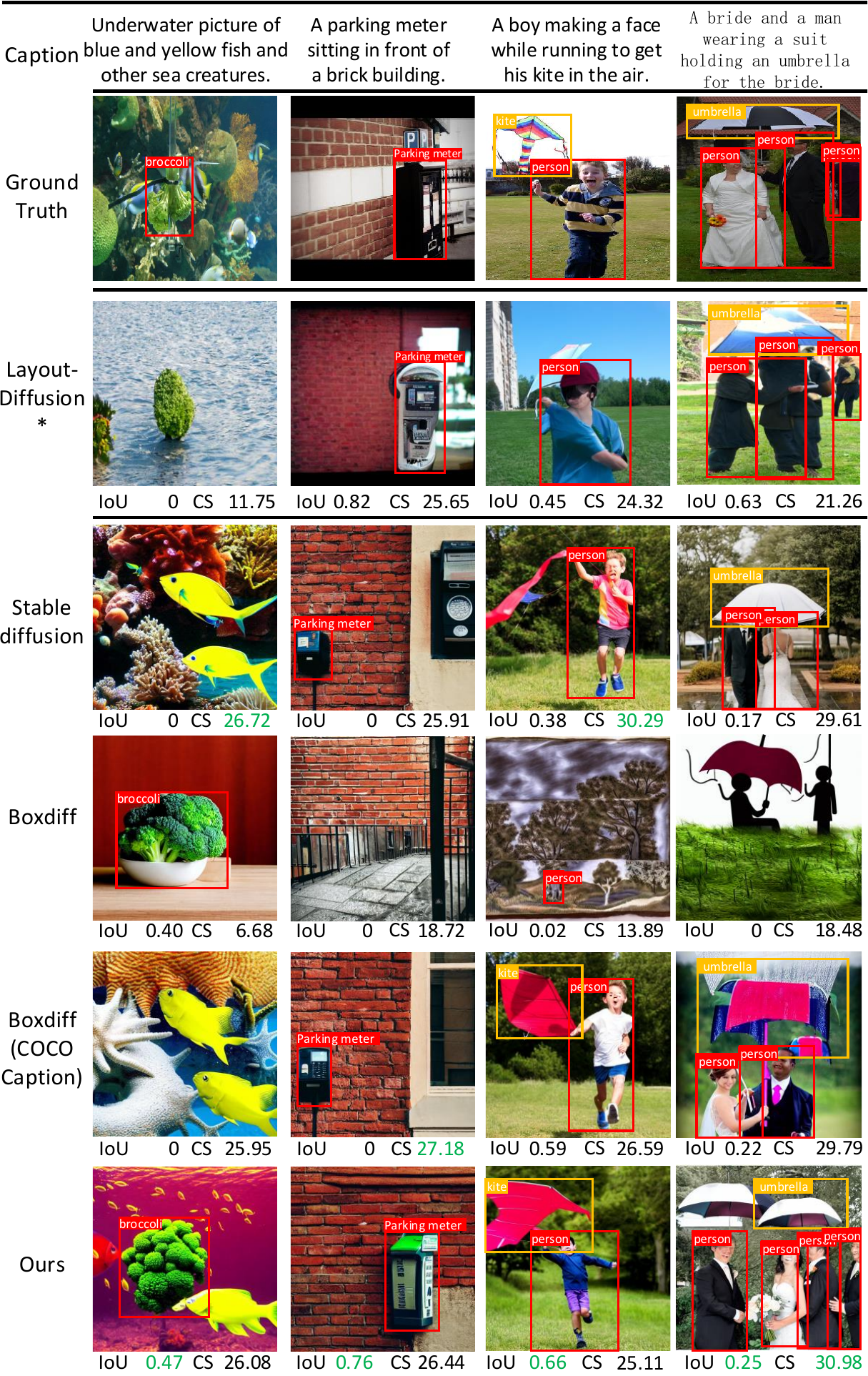}
    \caption{Images generated by different methods. We also report the IoU and CS score for each generated image. 
    %IoU and CS shows the accuracy of object localization and the relevance to the input text of generated images.
    % *: Requires training
    } % (Intersection over Union)  (CLIP Similarity)
    \label{copmarison_0}
\end{figure}

\subsection{Main Results}
Firstly and foremost, we report the overall performance of our proposed method on Coco-stuff dataset and compared with existing methods. As can be found in Tab.~\ref{tab:main}, our method outperforms existing SOTA, or BoxDiff~\cite{xie2023boxdiff}, significantly over all evaluation metrics. Such as averaged 22$\%$ relative improvement in terms of AP, AP50 and AP75. Interestingly, it is noticeable that when fed with simplified prompt, BoxDiff can hardly follow spatial constraints, whose score is very much close to the original stable diffusion models~\cite{rombach2022ldms}, while gives the worst FID score among all methods. This is understandable as the simplified prompts lack the description for contexts. With our revision on BoxDiff, its performances improves to some extent, or from $0.009$ to $0.0027$ and $60.75$ to $33.59$ in terms of AP and FID. This observation further demonstrates the difficulty of our goal. 

We also list the number of LayoutDiffusion~\cite{zheng2023layoutdiffusion} for completeness. Clearly, it gives the best performance among all methods. For instance, it is 8 and 10 times better than ours or existing SOTA method on AP. Nevertheless, our line of work is training-free and can be applicable to unseen objects at test time. We provide some visual examples of all methods in Fig.~\ref{copmarison_0}.

\begin{figure}[t]
    \centering
    \includegraphics[width=0.93\columnwidth]{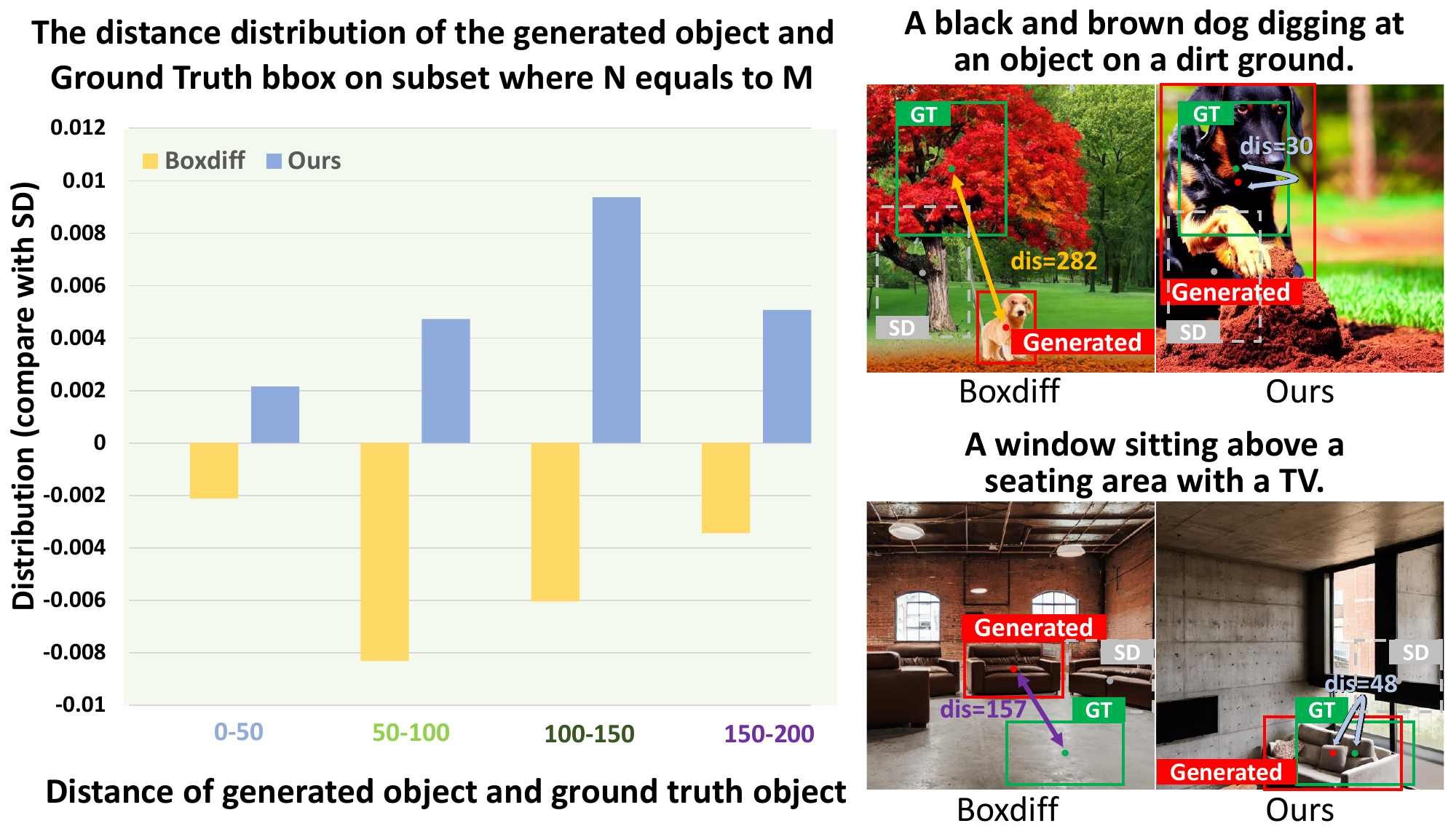}
    \caption{In the subset where N equals to M, our method is able to generate objects closer to the Ground Truth.}
    \label{subset_fig}
\end{figure}

\noindent{\textbf{Does geometric transformation relocate objects better?}} To better evaluate the effectiveness of our geometric transformation module, we conduct new experiments with a new subset. Specifically, we choose images where at most one object instance is annotated in spatial condition for one category, or $N$ equals to $M$. And then we follow the Boxdiff~\cite{xie2023boxdiff} to generate simplified prompt for each image. With the above mentioned criteria, we are able to remove the negative impact of semantic inconsistency in dataset and the fight between instance-level loss in BoxDiff~\cite{xie2023boxdiff}. We report the overall performance on the new subset, or 2153 images, in Tab.~\ref{tab:subset} and Fig.~\ref{subset_fig}. Please note that since different semantics might initialized at various locations, we also report the performance of stable diffusion as a reference, or starting points. Not surprisingly, our method significantly minimizes the distance between the center of generated objects and their ground truth, compared to SOTA BoxDiff. This observation clearly supports our claim on geometric transformation.

% \begin{table}[htbp]
%   \centering
%   \input{table/Ablation_Studies}
%   \caption{Ablation Studies}
%   \label{tab:ablation}
% \end{table}
\subsection{Ablation Studies}
To validate the effectiveness of our proposed method, we conduct through ablation studies on adding or removing any of the semantic and geometric constraints. Again, we report our experiments on Coco-stuff validation set and provide quantitative results in Tab.~\ref{tab:main}. There are several interesting observations. First of all, introducing each component will bring performance improvement, demonstrating the effectiveness of individual design. Secondly, the geometric transformation module is the key factor to improve the overall performance, indicating that explicitly manipulation is of great importance and can be a complimentary actuation for enforcing losses on attention map. Finally, it is clear that our full method gives the best performance, which supports our claim that these components are mutually effective.

\begin{table}[t]
  \centering
  
\begin{tabular}{cc|ccc} %{p{1cm}p{0.75cm}|p{1.5cm}p{1.5cm}c{1.5cm}}
\toprule
                                                                                   % &      & Stable\newline Diffusion & Boxdiff\newline(coco caption) & ours \\ \midrule
\makecell{Dataset}                                                                 &      & \makecell{Stable\\Diffusion} & \makecell{Boxdiff \\ (coco caption)}               & ours \\ \midrule
\multirow{3}{*}{\begin{tabular}[c]{@{}c@{}}N == M\\ subset\end{tabular}}           & AP   & 0.9              & 4.6                   & 5.1  \\
                                                                                   & AP50 & 3.2              & 13.1                  & 14.0 \\
                                                                                   & AP75 & 0.4              & 2.8                   & 3.0  \\ \hline
\multirow{3}{*}{\begin{tabular}[c]{@{}c@{}}prompt\\ editing\\ subset\end{tabular}} & AP   & 0.7              & 3.2                   & 3.8  \\
                                                                                   & AP50 & 2.4              & 8.7                   & 9.7  \\
                                                                                   & AP75 & 0.4              & 1.4                   & 2.2  \\ \bottomrule
\end{tabular}
  \caption{Ablations on semantic and geometric consistency. 
  %Results on proposed subsets shows both Semantic Consistency and Geometric Consistency are important. 
  }
  \label{tab:subset}
\end{table}

\noindent{\textbf{Is semantic consistency really necessary?}} We conduct one more experiment on semantic consistency. In this case, we collect images where prompt editing is necessary, and report the overall performance on them. Results in Tab.~\ref{tab:ablation} show that the performance of BoxDiff decreases on these images, leading to larger gaps when compared to our method.

\noindent{\textbf{Is geometric consistency as important?}} Besides semantic consistency, we also demonstrate the effectiveness of geometric consistency. Similar to the setup above, we construct a subset by selecting images as long as $N$ equals to $M$. Then we report the overall performance of baselines in Tab.~\ref{tab:ablation}. Again, ours gives the best performance, proving our concept of the importance of geometric consistency. 

\begin{figure}[htbp]
    \centering
    \includegraphics[width=0.93\columnwidth]{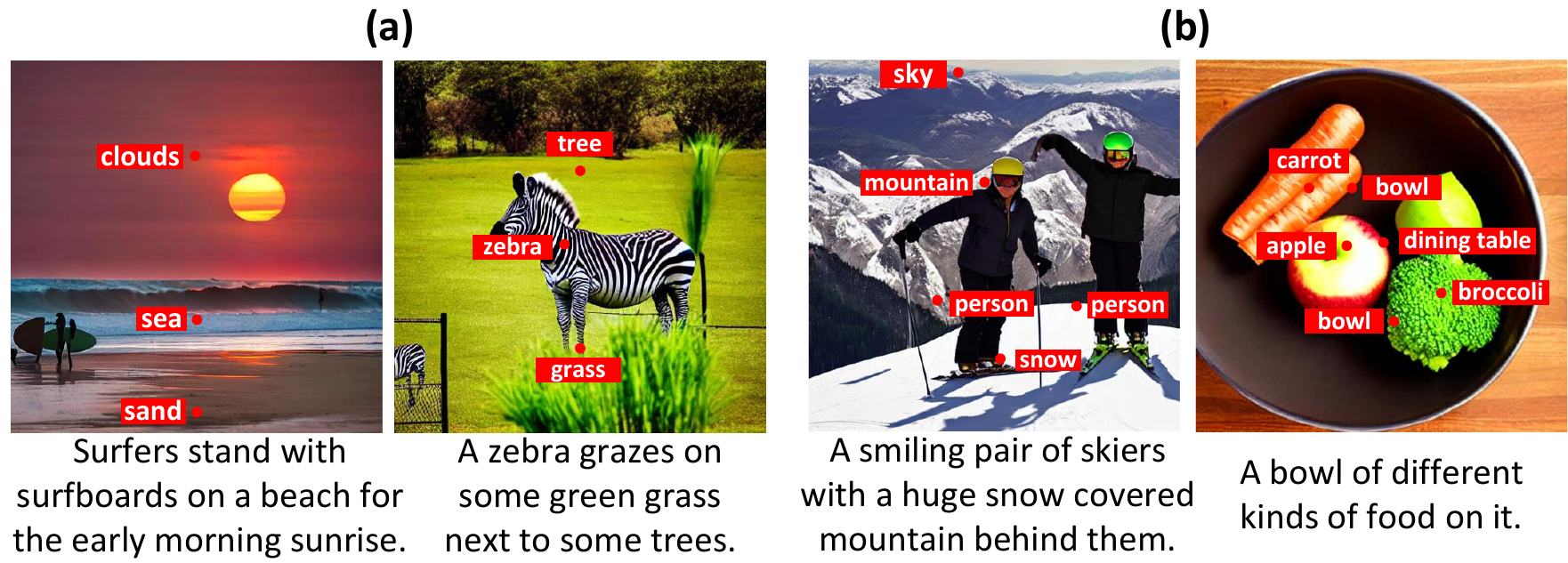}%Point_image_v3_2.pdf %空间不够则用v5
    \caption{Images generated by our method using keypoints(red point in the image). Our method has well performs under simple (a) and complex (b) conditions.}
    \label{point_image}
\end{figure}

\subsection{Extension to Diverse Spatial Constraints}
We further conduct simple extensions of our proposed method with diverse spatial constraints. Besides bounding boxes, we propose to relax the constraints by providing only one keypoint for each instance, mimicking the interface when users are asked to click on the image. 

In practice, we can refer to either existing dataset, such as training images in Coco-stuff~\cite{caesar2018cocostuff}, or ChatGPT~\cite{openai2023chatgpt4} to obtain the category-wise size information of target objects. Once converting these keypoints to bounding boxes with an assumption that a keypoint represents the center of a target bounding box, one can directly apply our method in point-wise setup. %We visualize some of our examples in Fig.~\ref{point_image}. 
Not surprisingly, our method can generate rather realistic images with diverse spatial constraints (See Fig.~\ref{point_image}).

\section{Conclusion}
\label{sec:conclusion}
In this paper, we propose a novel generic training-free image generation method that handles natural language prompts and complex spatial conditions. By decomposing the spatial conditions into semantic and geometric conditions, our method address the semantic-mismatch and geometric transformation problem in attention map and latents. respectively. 
%Specifically, we propose a novel geometric transformation module to explicitly manipulate on latents, leading to more precise relocation and refill during de-noising process. 
Experiments on Coco-stuff demonstrate the effectiveness of our proposed method over existing methods, showcasing our ability of coping with complex yet realistic scenarios.
% , as well as with diverse constraints.

\bibliography{main} %,aaai25
\newpage
\appendix
\label{supplementary}

% \title{supplementary}
% \begin{document}

% \maketitle

% \section{supplementary}

\section{Societal impact and existing limitations}
\begin{figure}[htbp]
    \centering %表示居中
    \includegraphics[width=\columnwidth]{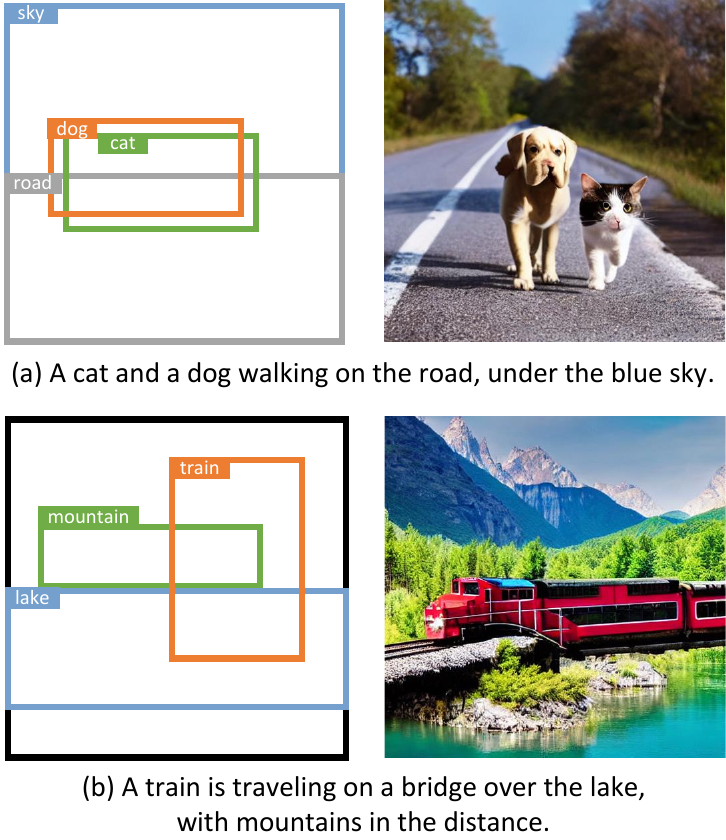} %height=6cm,
    \caption{The examples demonstrate the limitations of the current text-to-image model, we cannot generate the cat and the dog highly overlapping in (a) or a vertical train in (b).}
    \label{limitations_image}
\end{figure}

Our work has improved the controllability of text to image generative models, which helping users better control the layout during generating images, and making it easier for non-expert users to obtain the desired images. Improving the controllability of image generation enables image generation models to be better used for various tasks, such as 3D generation and comic generation.
However, there are still many limitations to the current image generation models. Here we provide some cases in which our method may fail to generate images in line with spatial conditions:\quad i) Objects with a high overlap ratio. (See Fig.~\ref{limitations_image} (a) )\quad ii) Objects with unusual shapes or sizes. (See Fig.~\ref{limitations_image} (b) )
%\ \ iii) Objects with unusual relationships.

\section{Dataset Construction}
We conduct our experiments on COCO 2017 Stuff  ~\cite{caesar2018cocostuff} and two newly proposed subsets. 

\textbf{COCO 2017 Stuff} dataset contains images bounding boxes and pixel-level segmentation masks for 80 categories of thing and 91 categories of stuff. Following the settings of LayoutDiffusion ~\cite{zheng2023layoutdiffusion}, we use the COCO 2017 Stuff Segmentation Challenge subset that contains 40K / 5k / 5k images for train / val / test-dev set and pick out 3,097 val images from the val set which satisfy: i) with 3 to 8 objects\ \ ii) cover more than 2\% of the image \ \ iii) not belong to crowd \ \ as our val dataset.

\textbf{prompt editing subset} is proposed to demonstrate the effectiveness of semantic consistency. We  pick out cases of prompts that have been modified during the generation process of our method as a new subset, including the cases with objects that did not included in the prompt and those with multiple objects of the same category.

\textbf{N == M subset} is proposed to demonstrate the effectiveness of geometric consistency. As described in the main text, N represents the number of objects, and M represents the number of object categories. We notice that the prompt editing will work when N \textgreater\  M. Therefore, to identify the images been generated with the effect of geometric consistency, we selected images where at most one object instance is annotated per category under spatial conditions to demonstrate the effectiveness of geometric consistency.
%We notice that the prompt editing is effective when N \textgreater\ M.

\section{Additional implementation details}

\begin{figure*}[tb]
    \centering %表示居中
    \includegraphics[width=0.95\textwidth]{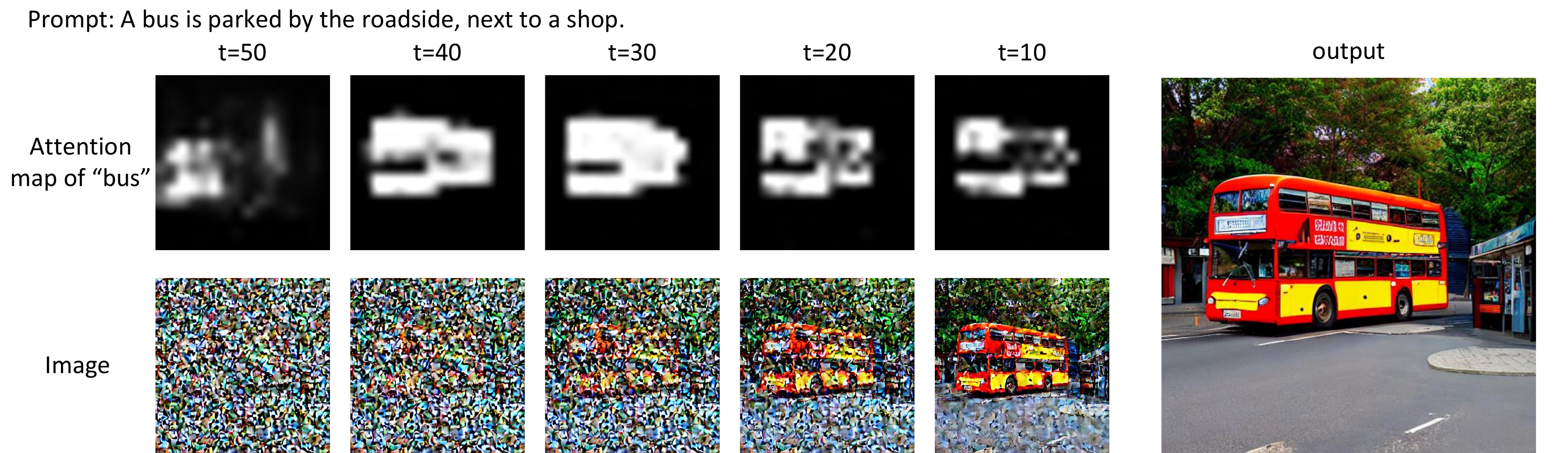} %height=6cm,
    \caption{We visualized attention maps and found that step $t=40$ is a good choice for performing prompt editing to rematch attention maps.}
    \label{attention_map_step}
\end{figure*}

\subsection{Diffusion model}
We use publicly available pretrained stable diffusion v1.4 as our backbone model, it consists of a denoising UNet to execute the denoising process within a compressed latent space and a VAE to connect the image and latent spaces. The pre-trained VAE of the Stable Diffusion is maintained with official weights and is used to encode images during the training phase and decode the latent codes into images during the inference phase. 
For all experiments, we use use 50 steps of DDIM sampling with a classifier-free guidance scale of 7.5 and attention map resolution of 16. With each prompt, our method will generate a RGB image $I$ whose resolution is 512 by 512. Attention can be performed at different scales, including 64$\times$64, 32$\times$32, 16$\times$16, and 8$\times$8. And we follow the convention to choose the scale of 16$\times$16. Latents are of the size 64$\times$64 since spatial details are largely remained at this scale compared to the low resolution ones. 
%ToDo

\subsection{Semantic Consistency details}
For all experiments, we perform Prompt editing after the input of prompt $l$ and spatial condition $c$. According to the visualization of attention maps (see Fig.~\ref{attention_map_step}), we observe that the attention maps generated during the initial stages of DDIM are sufficiently informative to discern whether the word is explicitly depicted in the target image to be generated, so we perform the attention map matching at $t=40$ of DDIM.

\subsection{Geometric Consistency details}
We perform Geometric Consistency in step 25 of DDIM, Here, we provide more details about our Geometric Consistency.

\begin{figure}[tb]
    \centering %表示居中
    \includegraphics[width=\columnwidth]{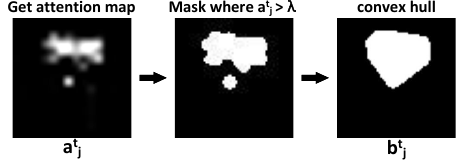} %height=6cm,
    \caption{We collected attention map and mask the object area, then convert the mask area to a convex hull.}
    \label{RoI_identification}
\end{figure}

\textbf{RoI identification} is proposed to identify the region-of-Interests in latents for $g_n$. The details after get attention map is visualized as the process in Fig.~\ref{RoI_identification}: for every $s_n$ we find the set of pixel greater than $\lambda$ in their corresponding attention map (as the middle of Fig.~\ref{RoI_identification}). Considering that objects are typically appear as continuous convex polygons in images, we use Andrew Algorithm %(See Alg.~\ref{alg:Andrew}) 
to get the convex hull of the set of pixels as our final RoI of $s_n$.

\begin{figure}[tb]
    \centering %表示居中
    \includegraphics[width=\columnwidth]{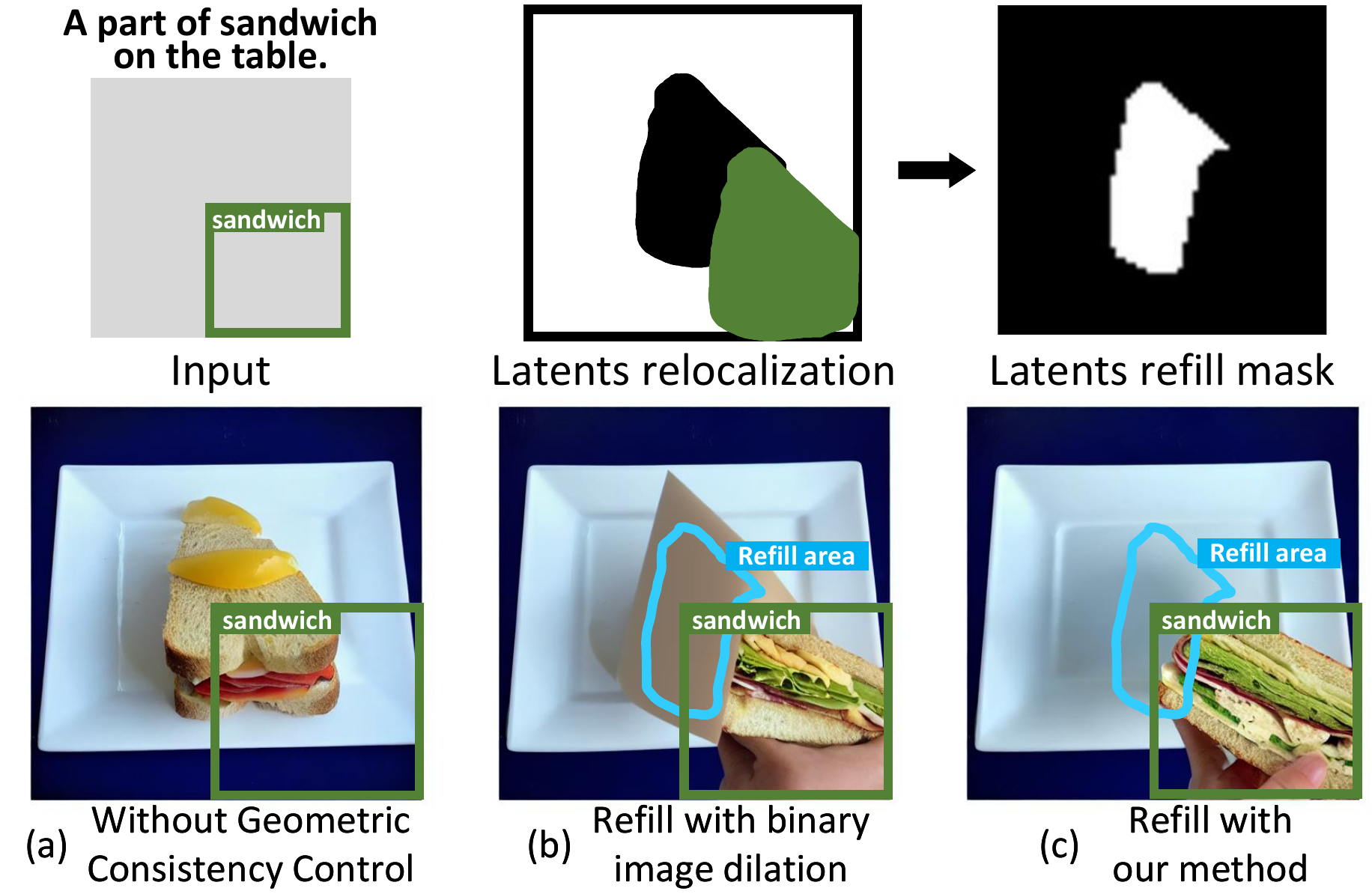} %height=6cm,
    \caption{Different methods in RoI refill. Our method(c) has fewer artifacts than using binary image dilation(b).}
    \label{RoI_refill_image}
\end{figure}

\textbf{RoI refill} is proposed to better fill the missing parts $b^t_n$ caused by RoI relocalization (see Fig.~\ref{RoI_refill_image}).  We first considered binary image dilation, an algorithm that can refill the missing areas conveniently. However, as image (b) shown in Fig.~\ref{RoI_refill_image}, the newly refilled areas exhibit significant distortion, resulting in a substantial degradation of the image quality we generate. As image (c) shown in Fig.~\ref{RoI_refill_image}, the proposed diffusion-based technique that learns to refill the latents, as described in the main text, addresses the previously mentioned issues.

\subsection{Computation of metrics}
We employed various methods to evaluate the quality of our generated images, as described in the main text. Here, we provide more details:

\begin{itemize} %Similar to 
\item \textbf{AP}. Following ~\cite{zheng2023layoutdiffusion}, in our experiment, We utilize the publicly available YOLOv4 ~\cite{bochkovskiy2020yolov4}, which is trained on the COCO 2017 Stuff dataset ~\cite{caesar2018cocostuff}, to detect 80 different categories of objects in the generated images, providing bounding boxes for each detected object. We regard the spatial conditions as the ground truth and compare them with the detection results of the generated images, compute the AP values using the COCOEval module of pycocotools Python library.

\item \textbf{CS} (CLIP Score ~\cite{radford2021learning}). In our experiment, we employ a metric of CLIP Score to explicitly evaluate the correctness of semantics in the synthesized images. Specifically, we use the publicly available pretrained CLIP ViT-L/14 model to process both images and text, and then calculate their similarity.

\item \textbf{FID} (Fr‘echet Inception Distance ~\cite{heusel2017fid}) shows the difference between the real images and the generated images by using a pretrained Inception-V3 ~\cite{SzegedyVISW16} network. We use images in COCO dataset as ground truth to evaluate the generated images. The FID score is computed using the official FID code, with the dataset folder and the generated image folder as inputs. A lower FID score suggests that the generated images are more similar to the real images, indicating a better quality of the generated images.

\end{itemize}

\subsection{Details of computing the distance distribution}
To verify that our method can bring the generated objects closer to the ground truth, we provide the distance distribution (compared with stable diffusion) in the experiment of the proposed subset where N equals to M. To compute the distance distribution (compared with stable diffusion), for $n_{gt}$ items in the ground truth, we first compute their Euclidean distance to their matching generated objects. Then, for every distance interval $k$, we count the number of objects in the distance interval and compute the probability as following Eq.~\ref{eq:compute_dis_1}.
\begin{equation}
\begin{split}
  \textbf{dis} &= \sqrt {(x_0-x)^2+(y_0-y)^2} \\ 
  &\textbf{P}_{k} = \frac{1}{n_{gt}} \sum_{i=1}^{n_{gt}} I\ _{ k }
\end{split}
\label{eq:compute_dis_1}
\end{equation}
Here, the distance interval $k$ is of the form $ k : 0 \leq dis_i \leq 50$, $I$ is an indicator function that takes the value 1 when $dis_i$ is within the distance interval $k$, and 0 otherwise.
\begin{equation}
\begin{split}
  \Delta \textbf{P}_{k}^{Boxdiff} &= \textbf{P}_{k}^{Boxdiff}-\textbf{P}_{k}^{SD} \\ 
  \Delta \textbf{P}_{k}^{Ours} &= \textbf{P}_{k}^{Ours}-\textbf{P}_{k}^{SD}
\end{split}
\label{eq:compute_dis_2}
\end{equation}
As shown in Eq.~\ref{eq:compute_dis_2}, we compute the probability of the distance of generated objects by Boxdiff and Ours for each distance interval $k$ separately, and then subtract the results from those of stable diffusion to obtain the final result.

\section{Additional experiments}
\begin{figure}[htbp]
    \centering %表示居中
    \includegraphics[width=\columnwidth]{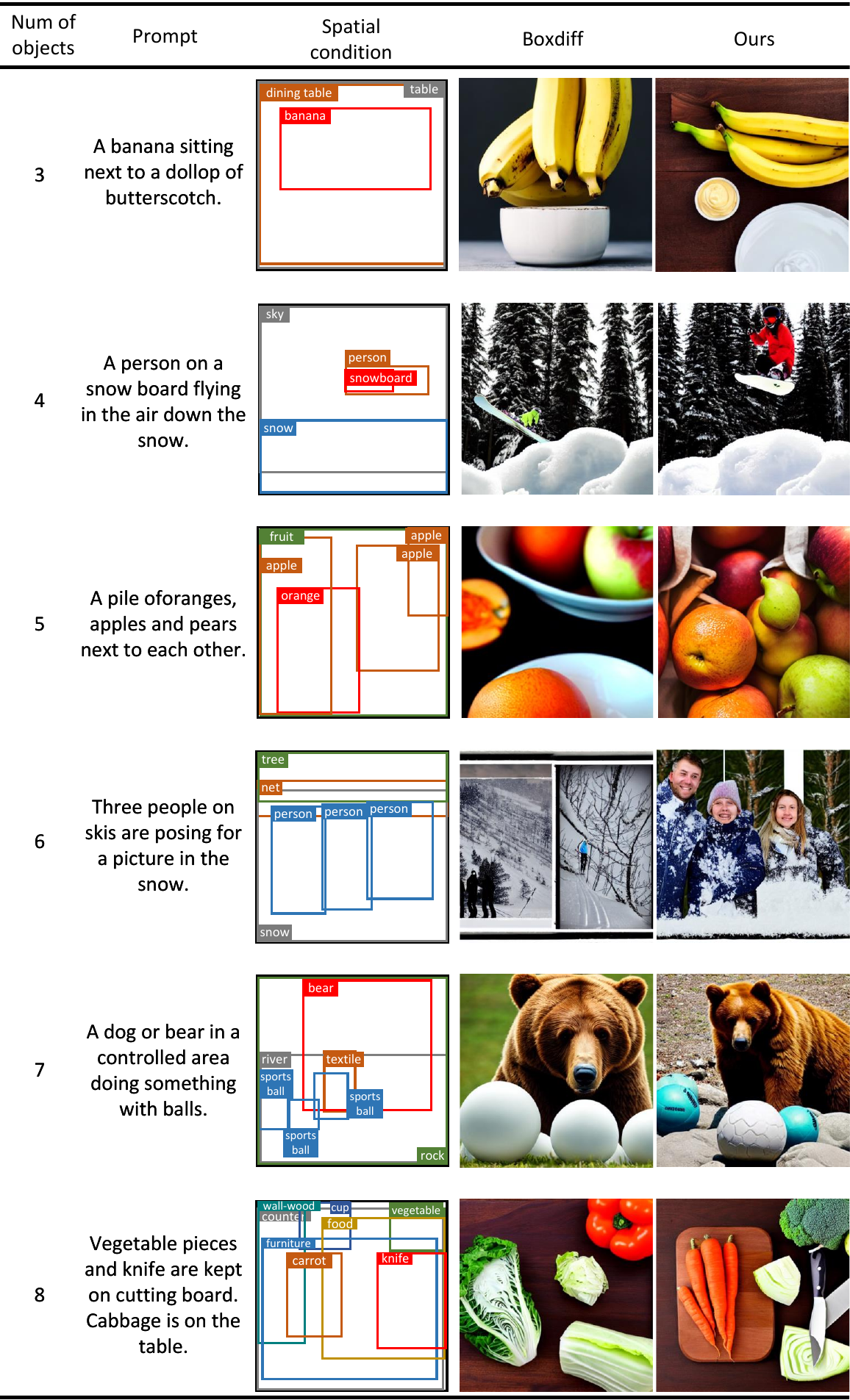} %height=6cm,
    \caption{Examples of spatial conditions with varying numbers of objects in the COCO dataset. The visualized generated image shows that our method generates more accurate object positions and fewer missing objects.}
    \label{fig:num_obj}
\end{figure}

% AP for different num of objs
% \subsection{Performance on different quantities of objects}
% In order to observe the performance of the proposed method under input conditions containing different numbers of objects. We classified the instances in the dataset into 6 groups based on the number of objects in each image, and report the performance of the proposed method for each group. 
To evaluate the performance of the proposed method across varying object quantities in input conditions, we categorized dataset instances into six groups based on the number of objects per image, and presented the method's performance for each group.
As can be found in Tab.~\ref{tab:num_of_objs}, our method has an average relative improvement rate of 20\% in AP across different quantities of objects, compared to the existing SOTA Boxdiff~\cite{xie2023boxdiff}. The improvement rates are 17.4\% and 22.8\% on AP50 and AP75, respectively. From this, it can be seen that our method has significant advantages in layout consistency, especially when there are more objects, our method has a greater improvement compared to Boxdiff. Here we provide some examples in Fig.~\ref{fig:num_obj}.

% \subsection{Make generated items and ground truth closer}
% As described in the main text, we 
% visualize distance for fig9

\begin{table}[htbp]
  \centering
  
\begin{tabular}{c|ccc|ccc}
\toprule
\multirow{2}{*}{\begin{tabular}[c]{@{}c@{}}Num of\\ objects\end{tabular}} & \multicolumn{3}{c|}{Boxdiff} & \multicolumn{3}{c}{Ours} \\ 
%\cline{2-4} \cline{5-7}
                                                                          & AP     & AP50     & AP75    & AP    & AP50    & AP75   \\ \midrule
3                                                                         &  9.8   & 21.1     &  9.2    & 10.0  & 24.2    &  8.6   \\
4                                                                         &  7.6   & 17.8     &  6.5    &  7.1  & 16.7    &  4.9    \\
5                                                                         &  4.4   & 12.8     &  2.5    &  6.0  & 15.0    &  3.7    \\
6                                                                         &  4.1   & 10.5     &  2.3    &  4.5  & 12.4    &  2.4    \\
7                                                                         &  3.1   &  7.8     &  1.9    &  3.8  & 10.2    &  2.2    \\
8                                                                         &  2.5   &  7.9     &  0.7    &  3.9  & 10.2    &  1.4    \\ \bottomrule
\end{tabular}
  \caption{Different quantities of objects per image. Our method performs better with images containing different quantities of objects, especially when there are many objects.} % counts or quantities
  \label{tab:num_of_objs}
\end{table}

\begin{figure*}[t]
    \centering %表示居中
    \includegraphics[width=0.98\linewidth]{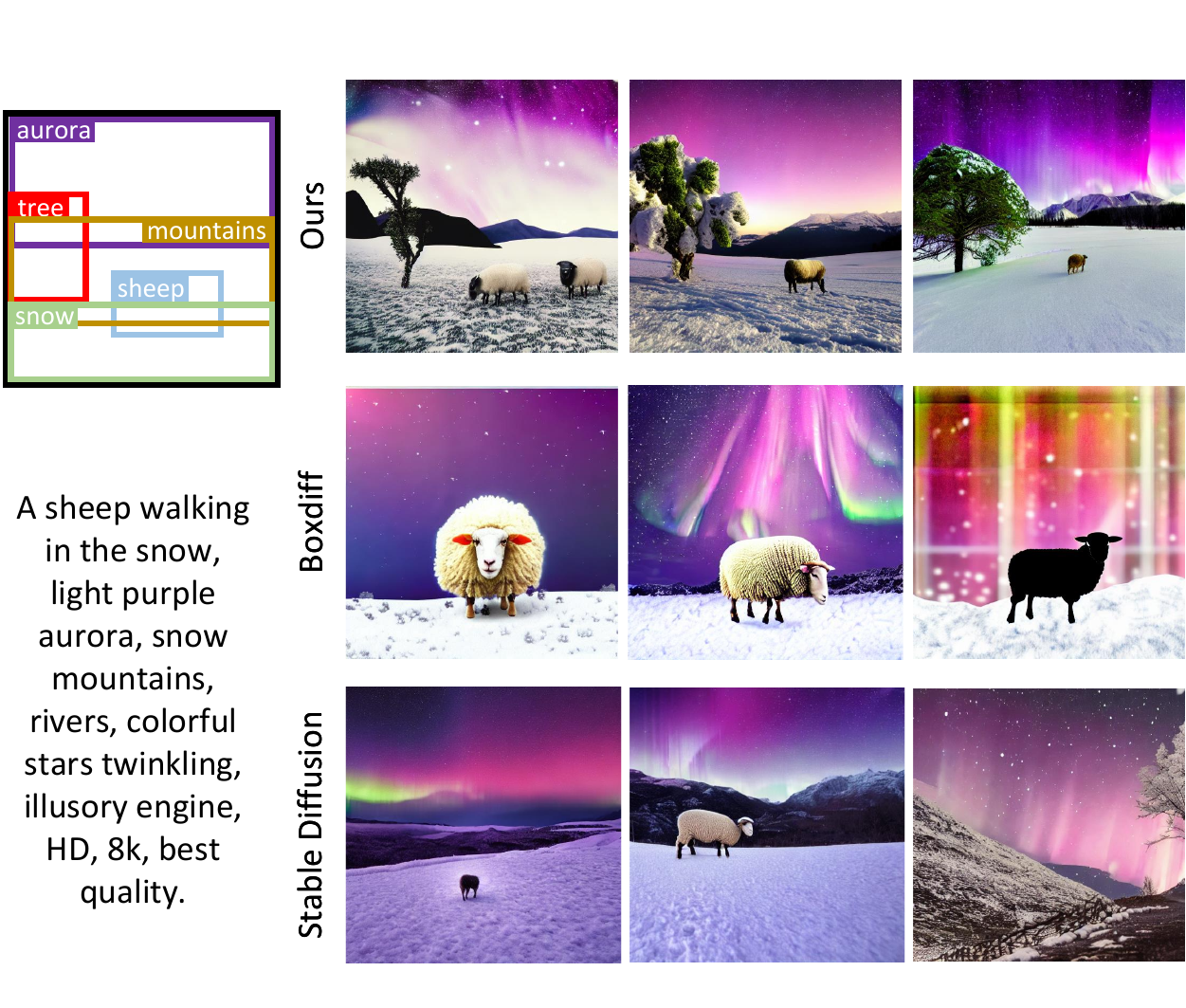} %height=6.5cm,  width=16cm
    \caption{Camparing with Boxdiff and Stable Diffusion, our method is able to generates objects such as ``tree" and ``mountains" in the correct position according to spatial conditions.}
    \label{vis_3}
\end{figure*}

\section{Additional visualizations}
\begin{figure*}
    \centering %表示居中
    \includegraphics[width=0.9\linewidth]{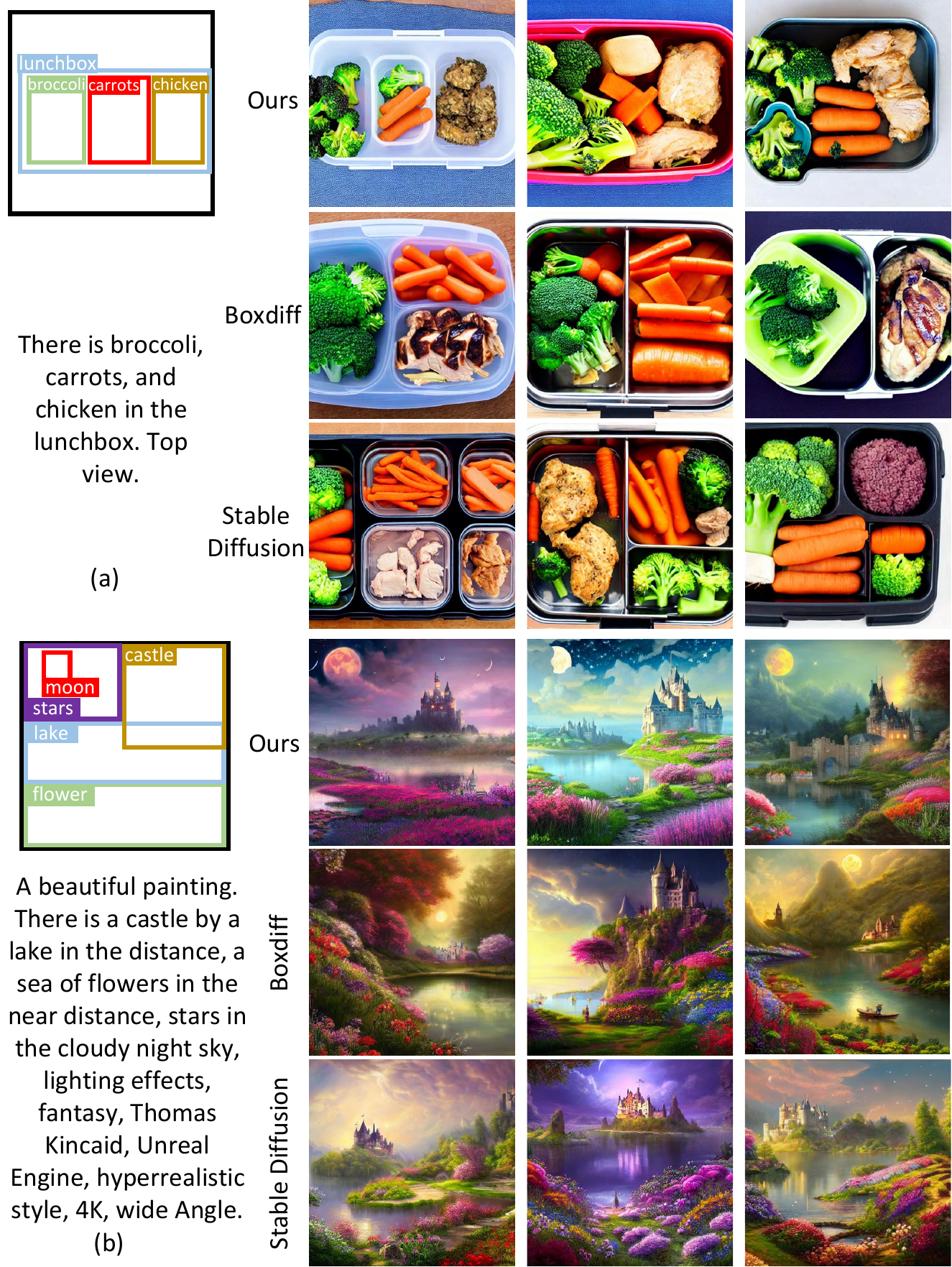} %height=6.5cm,  width=16cm
    \caption{In example (a), our proposed method correctly generate three kinds of food in the lunchbox according to the spatial conditions. In example (b), our method can generate multiple objects according to the spatial conditions in a complex scene without missing any objects.}
    \label{vis_4}
\end{figure*}

\begin{figure*}
    \centering %表示居中
    \includegraphics[width=0.9\linewidth]{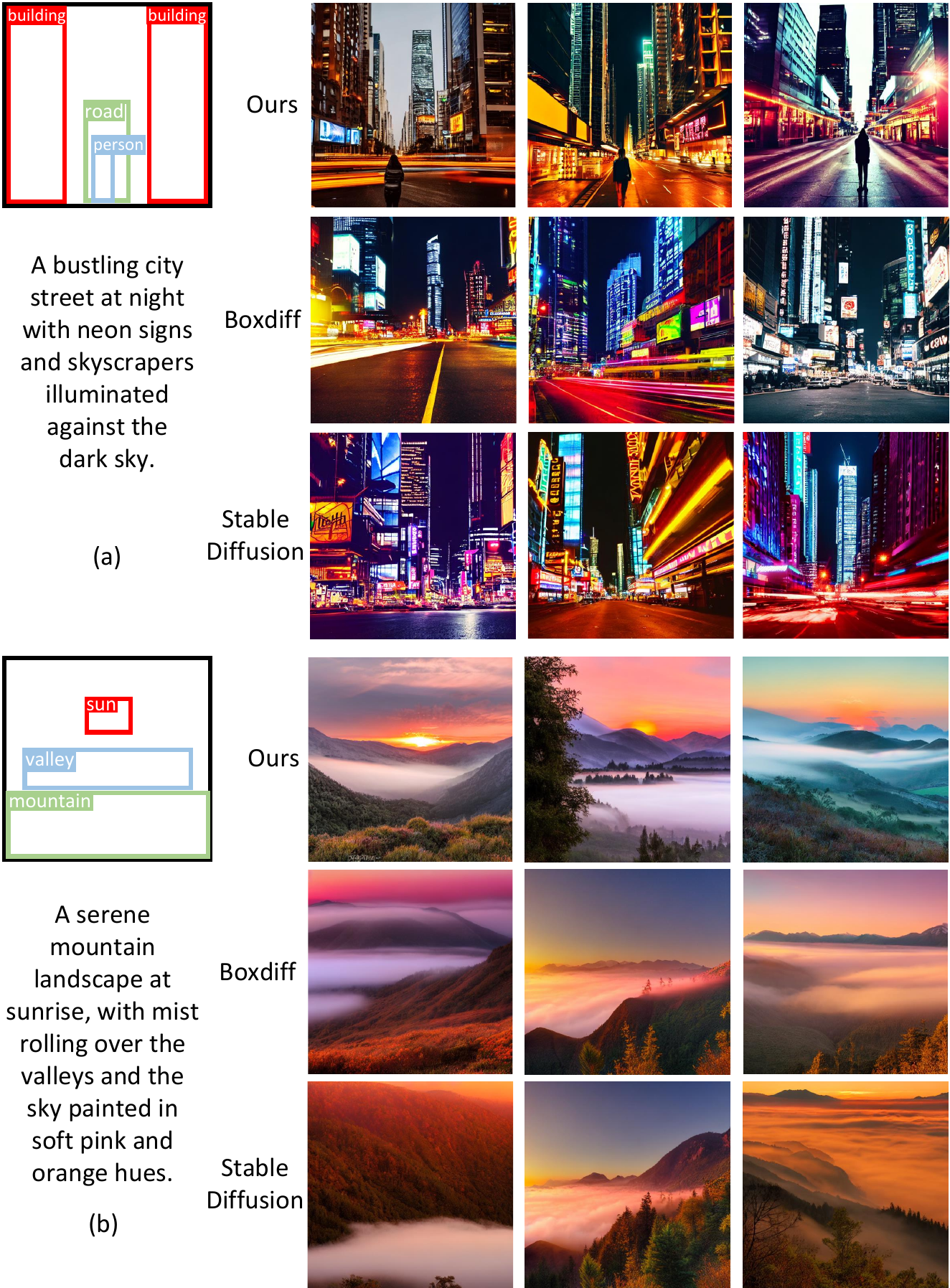} %height=6.5cm,  width=16cm
    \caption{The spatial condition ``person" in example (a) and the spatial condition ``sun" in example (b) are missing in methods other than ours.}
    \label{vis_1}
\end{figure*}

\begin{figure*}
    \centering %表示居中
    \includegraphics[width=0.9\linewidth]{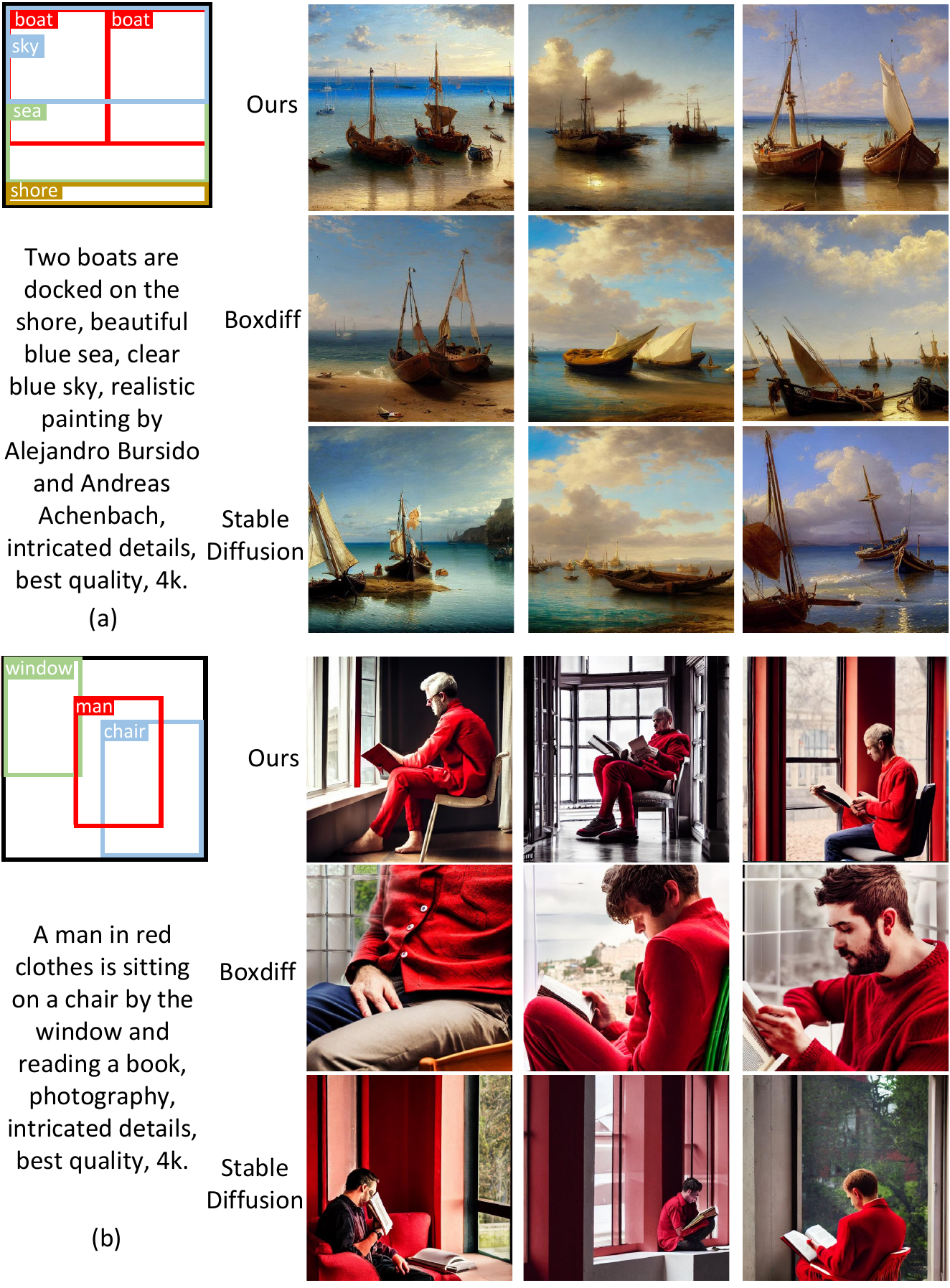} %height=6.5cm,  width=16cm
    \caption{In example (a), our method correctly generated two ``boats" in the left and right of the image. In example (b), the image generated by our method correctly positions three objects while also meeting semantic conditions.}
    \label{vis_2}
\end{figure*}

\begin{figure*}
    \centering %表示居中
    \includegraphics[width=\linewidth]{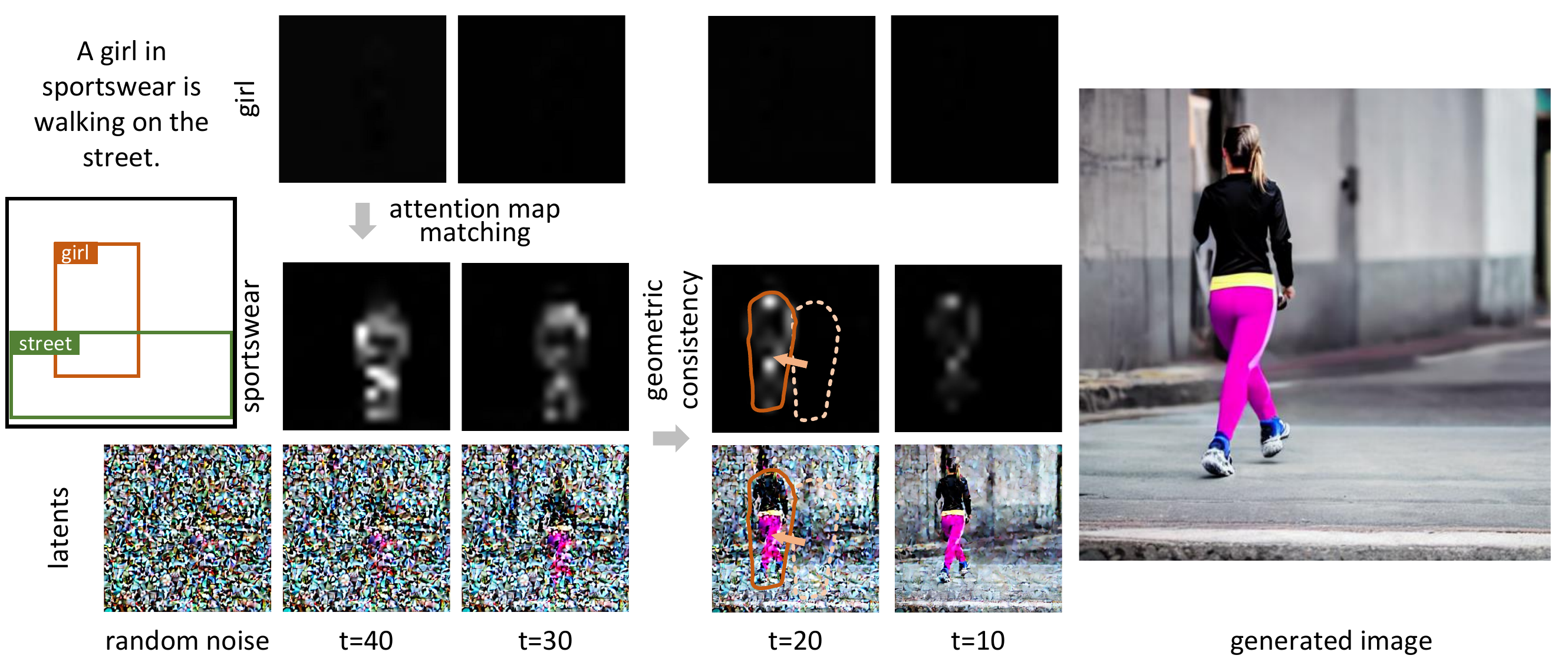} %height=6.5cm,  width=16cm
    \caption{Semantic consistency matches the correct attention map for the ``girl", and geometric consistency ``move" the ``girl" to the correct position.}
    \label{vis_method1}
\end{figure*}

\begin{figure*}
    \centering %表示居中
    \includegraphics[width=\linewidth]{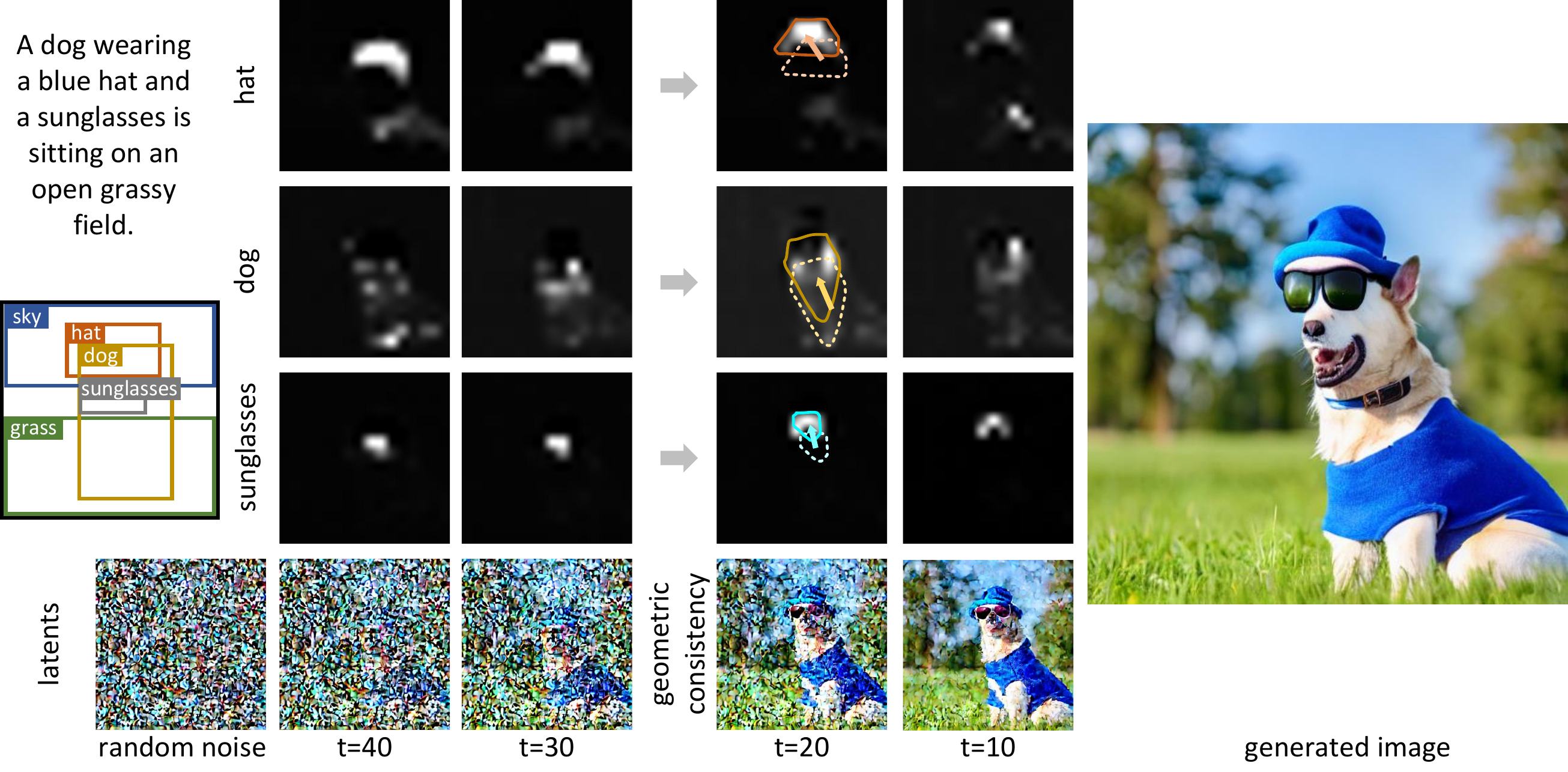} %height=6.5cm,  width=16cm
    \caption{The ``hat", ``dog" and ``sunglasses" are relocalized by geometric consistency and been generated in correct position.}
    \label{vis_method2}
\end{figure*}

In this section, we provide more visual comparison among Stable Diffusion~\cite{rombach2022high}, Boxdiff~\cite{xie2023boxdiff} and our method (See Fig.~\ref{vis_3}, Fig.~\ref{vis_4}, Fig.~\ref{vis_1} and Fig.~\ref{vis_2}). It can be seen that the images generated by stable diffusion and Boxdiff may have missing objects like sun in Fig.~\ref{vis_1} (b) fig or inaccurate positions of objects like person in Fig.~\ref{vis_2} (a). Conversely, with the effectiveness of semantic consistency and geometric consistency, objects in the image generated using our proposed method are positioned with greater accuracy. \\
To more clearly demonstrate the actual effects of semantic consistency and geometric consistency in controlling the layout of objects during the generation process, we visualized some examples of generation process in Fig.~\ref{vis_method1} and Fig.~\ref{vis_method2}, and marked the positions where the effects of semantic consistency and geometric consistency were applied during the generation process. In Fig.~\ref{vis_method1}, our proposed method rematched the attention map and adjusted the position of the ``girl" correctly. In Fig.~\ref{vis_method2}, our method relocalized the ``dog", ``hat" and ``sunglasses", ensuring that their positions align more closely with the spatial conditions.\\
As can be seen, both semantic consistency and geometric consistency play significant roles during the generation process, leading to improved controllability in the images generated by our method. Note that all these visualized images are generated using our method or Boxdiff with the publicly available pretrained stable diffusion v1.4.

% \bibliography{main} %,aaai25

% \end{document}
% \input{supplementary}

\end{document}